\documentclass[11pt]{article}

\usepackage[final]{acl}

\usepackage{times}
\usepackage{latexsym}

\usepackage[T1]{fontenc}
\usepackage[utf8]{inputenc}

\usepackage{microtype}

\usepackage{inconsolata}

\usepackage{graphicx}

\usepackage{CJKutf8}
\usepackage{caption}   % 更好的 caption 格式
\usepackage{tabularx}
\usepackage{seqsplit}
\usepackage{booktabs}
\usepackage{multirow}
\usepackage{makecell}
\usepackage{amsmath}
\usepackage{amssymb}
\usepackage{tikz}
\usetikzlibrary{positioning}
\usepackage{adjustbox}
\usetikzlibrary{calc}
\usetikzlibrary{shadows, backgrounds, fit} % <<--- fit 库很关键！
\definecolor{myblue}{RGB}{219, 238, 244} % 淡蓝色，用于输入
\definecolor{myorange}{RGB}{252, 234, 220} % 淡橙色，用于高亮
\definecolor{mygreybg}{RGB}{245, 245, 245} % 浅灰色背景
\definecolor{mygreyline}{RGB}{180, 180, 180} % 灰色线条
\definecolor{mygreen}{RGB}{220, 240, 220} % 浅绿色，可以用于输出
\usepackage{algorithm}
\usepackage{algpseudocode}
\usepackage{placeins}

\title{MAAM: Anchor-Preserving Compression and Contextual Calibration for Chinese Discriminatory Language Detection}

\author{
  Yuxin Fu \and
  Shijing Si\thanks{Corresponding author.} \\
  School of Economics and Finance, Shanghai International Studies University \\
  Shanghai, China \\
  \texttt{yuxinfuNLP@outlook.com} \quad
  \texttt{shijing.si@shisu.edu.cn}
}

\begin{document}
\maketitle
\begin{abstract}
Chinese discriminatory-language detection is challenging because harmful intent is often implicit and context-dependent. We propose MAAM (Myopia--Astigmatism Anchor Mechanism), a lightweight, \textit{model-agnostic} framework inspired by functional visual blur: rather than preserving every token equally, MAAM retains discrimination-relevant semantic anchors and calibrates them with C--I--S contextual priors (Contextual Tone, Group Identity, and Stance Polarity). We also introduce ChLGBT, to our knowledge the first Chinese LGBT-focused discriminatory-language dataset, with 8,120 manually annotated samples and three ordinal labels: explicit bias, implicit bias, and emotional intensity. Across strong encoder baselines, MAAM improves all three prediction dimensions, with consistent gains in accuracy, F1, Brier score, and expected calibration error. Compared with frontier LLM baselines under zero-shot and few-shot prompting protocols, MAAM remains competitive while offering stronger compactness and stability. These results suggest that interpretable anchor preservation and contextual calibration provide a practical alternative to heavier model scaling for Chinese discriminatory-language assessment.
\end{abstract}

\section{Introduction}
Chinese discriminatory-language detection remains underexplored, and many existing methods depend on heavy model scaling \cite{zhao2023chbias, deng2022cold, chen2024chinese}, which is costly in practice \cite{strubell2019energy}. In this work, we challenge the full-input assumption and show that preserving semantic anchors under structured compression can improve discrimination-related judgments without enlarging the backbone model.

Human perception offers a useful analogy: early visual processing prioritizes structure over raw detail \cite{marr2010vision}. This ``functional blur'' view is relevant to high-context Chinese discourse \cite{hall1976beyond, gao1998communicating}, where discriminatory intent is often distributed across context rather than explicit keywords \cite{deng2022cold, elsherief2021latent, kim2022generalizable}. Prior abusive-language studies also show that reliable judgment depends on contextual information \cite{lopez2025context, bourgeade2024humans}. Evidence from vision and predictive-processing literature further supports structure-focused and context-aware inference \cite{geirhos2018imagenet, clark2013whatever}.

The terms Myopia and Astigmatism follow this perceptual analogy: mild visual blur can still preserve coarse structure, salient anchors, and contextual regularities. We transfer this intuition to language, where robust discriminatory-language detection may depend less on preserving every token than on retaining task-relevant anchors and contextual priors.

We instantiate this idea as \textbf{MAAM (Myopia--Astigmatism Anchor Mechanism)}, a lightweight, model-agnostic framework that combines anchor-preserving compression with contextual calibration for robust discriminatory-language assessment.

Existing Chinese bias/discrimination benchmarks such as CBBQ, ChBias, and Corgi-PM are mainly formulated with binary labels or coarse categorical targets \cite{huang2024cbbq,zhao2023chbias,zhang2023corgi}, limiting fine-grained analysis of discriminatory expression.
To address this gap and the scarcity of Chinese LGBT-focused resources, we formulate ChLGBT as three parallel ordinal prediction tasks: explicit bias, implicit bias, and emotional intensity, each annotated on a 1--5 scale.

Our contributions are threefold:
\begin{itemize}
    \item We propose MAAM, a plug-and-play framework that unifies anchor-preserving compression (Myopia) and C--I--S prior calibration (Astigmatism).
    \item We release ChLGBT, to our knowledge the first Chinese LGBT-focused discriminatory-language dataset, with 8,120 samples and three-dimensional annotations.
    \item We show that MAAM improves both fine-grained ChLGBT prediction and coarse-grained COLD Region/Race tasks, indicating cross-domain applicability beyond a single bias category while offering a practical alternative to model scaling under our evaluation protocol.
\end{itemize}
\begin{figure*}[t]
\centering
\includegraphics[width=\textwidth]{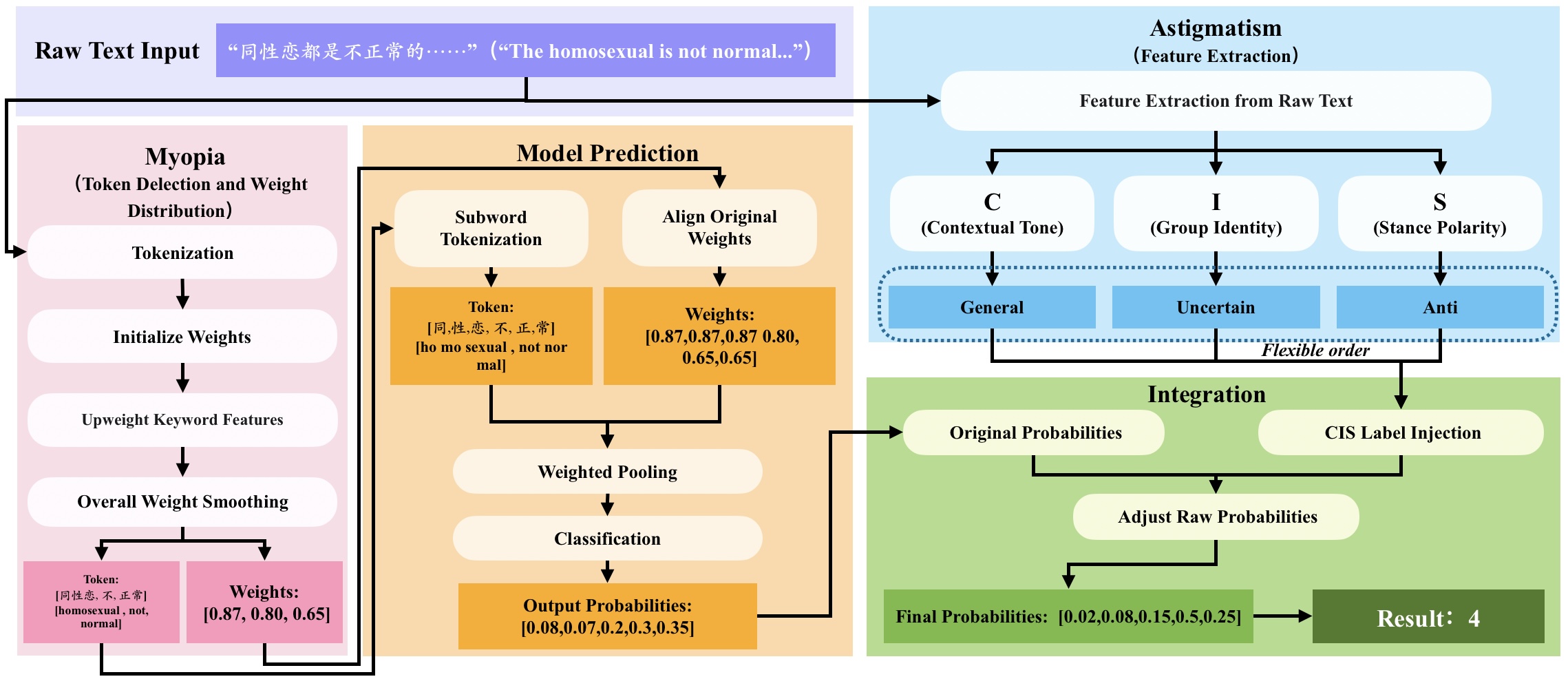}
\caption{MAAM framework (Explicit track shown as an example; Implicit and Emotional use the same pipeline).}
\label{fig:maam_pipeline}
\end{figure*}

\section{Related Work}
\subsection{Bias \& Hate Speech Detection}
Cross-lingual resources for hate/bias detection are growing \cite{cascalheira2024lgbtq+, chakravarthi2021dataset}, but Chinese LGBT-focused discriminatory-language datasets remain limited, and existing Chinese benchmarks mainly target broader social stereotypes. To address this resource gap, we release ChLGBT, a dedicated Chinese LGBT-focused dataset designed to support fine-grained analysis of explicit bias, implicit bias, and emotional intensity.

\subsection{Text Compression and Semantic Compensation}
Efficiency-oriented compression methods reduce token budgets \cite{goyal2020powerbert, kim2021ltp}, but can discard identity and pragmatic cues crucial for bias detection. MAAM's Myopia module is a task-aware compression strategy: it suppresses non-essential tokens while preserving bias-relevant semantic anchors. This differs from generic pruning and is closer to controlled semantic compensation \cite{leite2023noisy, xie2025simple}.

\subsection{Model Calibration and Prior Knowledge Fusion}
Neural classifiers are often overconfident, so robust bias detection also requires calibration \cite{guo2017calibration, kadavath2022language, ICLR2025_9a30247f}. Prior-guided calibration methods improve stability by injecting contextual signals at decision time. MAAM follows this line with Astigmatism, which fuses C--I--S priors with model outputs in a lightweight, architecture-agnostic way.
\begin{table*}[ht]
\centering
\scriptsize
\addtolength{\tabcolsep}{-4.5pt}
\begin{tabularx}{\textwidth}{@{} l l >{\raggedright\arraybackslash}p{2.8cm} X X ccc @{}}
\toprule
\textbf{Layer} & \textbf{Subcategory} & \textbf{Description} & \textbf{Example (EN)} & \textbf{Example (ZH)} & \textbf{Exp.} & \textbf{Imp.} & \textbf{Emo.} \\ \midrule
\multirow{5}{*}{\makecell[l]{1: Core\\Arguments}}
& Explicit Subject & Agent of action & ``community / netizens'' & \begin{CJK*}{UTF8}{gkai}“社区/网友”\end{CJK*} & 1 & 1 & 3 \\
& Explicit Object & Patient of action & ``same-sex marriage / LGBT people'' & \begin{CJK*}{UTF8}{gkai}“同性婚姻/同性恋群体”\end{CJK*} & 1 & 1 & 3 \\
& Implicit Subject & Unexpressed agent (e.g., passive) & ``(someone) required to come out'' & \begin{CJK*}{UTF8}{gkai}“（被）要求出柜”\end{CJK*} & 2 & 1 & 3 \\
& Pronouns & He, she, they, this, those & ``he / she / they / this / those'' & \begin{CJK*}{UTF8}{gkai}“他/她/它/他们/这个/那些”\end{CJK*} & 2 & 2 & 3 \\
& Existential Anchor (Entity) & Noun governed by ``exist/have'' & ``disagreement / bias'' & \begin{CJK*}{UTF8}{gkai}“分歧/偏见”\end{CJK*} & 4 & 4 & 4 \\ \midrule
\multirow{4}{*}{\makecell[l]{2: Attributive\\Modifiers}}
& Adjectival Modifiers & Descriptive adjectives & ``homophobic / discriminatory'' & \begin{CJK*}{UTF8}{gkai}“恐同的/歧视性的”\end{CJK*} & 1 & 1 & 1 \\
& Strong Polarity Intensifiers & Degree modifiers on evaluative words & ``extremely / highly'' & \begin{CJK*}{UTF8}{gkai}“非常/极其”\end{CJK*} & 2 & 3 & 1 \\
& Nominal Modifiers & Noun-noun compounds & ``lesbian students / gay community'' & \begin{CJK*}{UTF8}{gkai}“女同性恋学生/同性恋群体”\end{CJK*} & 2 & 2 & 4 \\
& Structural Particle ``de'' & Links modifier and head & ``de / di / de'' & \begin{CJK*}{UTF8}{gkai}“的/地/得”\end{CJK*} & 5 & 5 & 5 \\ \midrule
\multirow{3}{*}{\makecell[l]{3: Predicative\\Events}}
& Core Predicate Verb & Main action verb & ``support / oppose'' & \begin{CJK*}{UTF8}{gkai}“支持/反对”\end{CJK*} & 1 & 2 & 1 \\
& Equative Copula & Identity/classification judgments & ``is / are'' & \begin{CJK*}{UTF8}{gkai}“是”\end{CJK*} & 1 & 1 & 2 \\
& Existential Anchor (Dynamic) & ``Exist/have'' as root predicate & ``exist / have'' & \begin{CJK*}{UTF8}{gkai}“存在/有”\end{CJK*} & 4 & 4 & 4 \\ \midrule
\multirow{5}{*}{\makecell[l]{4: Modality\\ \& State}}
& Negation & Denial or prohibition & ``not / no'' & \begin{CJK*}{UTF8}{gkai}“不/没/无”\end{CJK*} & 1 & 1 & 1 \\
& Aspect Markers & Perfective/continuous aspect & ``le / zhe / guo'' & \begin{CJK*}{UTF8}{gkai}“了/着/过”\end{CJK*} & 5 & 3 & 4 \\
& Result Complements & Outcome-oriented complements & ``-de + harshly / clearly'' & \begin{CJK*}{UTF8}{gkai}“得很凶/得很清楚”\end{CJK*} & 3 & 3 & 1 \\
& Degree Complements & Extreme evaluation & ``-to the extreme / utterly'' & \begin{CJK*}{UTF8}{gkai}“透了/极了”\end{CJK*} & 3 & 3 & 1 \\
& Disjuncts / Parentheticals & Speaker-comment insertions & ``honestly / frankly'' & \begin{CJK*}{UTF8}{gkai}“说实话/坦白说”\end{CJK*} & 5 & 2 & 3 \\ \midrule
\multirow{5}{*}{\makecell[l]{5: Contextual\\Frames}}
& Locative Adjuncts & Spatial context & ``in China / on campus'' & \begin{CJK*}{UTF8}{gkai}“在中国/在校园里”\end{CJK*} & 4 & 2 & 3 \\
& Temporal Adjuncts & Time reference & ``recently / in recent years'' & \begin{CJK*}{UTF8}{gkai}“最近/近年来”\end{CJK*} & 4 & 3 & 3 \\
& Manner Adjuncts & How an action is performed & ``openly / privately'' & \begin{CJK*}{UTF8}{gkai}“公开地/私下地”\end{CJK*} & 3 & 1 & 2 \\
& Causal Adjuncts & Reason for action & ``because of religion / tradition'' & \begin{CJK*}{UTF8}{gkai}“因为宗教/因为传统”\end{CJK*} & 3 & 1 & 2 \\
& Purpose Adjuncts & Goal of action & ``for stability / for image'' & \begin{CJK*}{UTF8}{gkai}“为了稳定/为了形象”\end{CJK*} & 3 & 2 & 3 \\ \midrule
\multirow{4}{*}{\makecell[l]{6: Logical\\Connectors}}
& Adversative Conjunctions & Contrast & ``however / nevertheless'' & \begin{CJK*}{UTF8}{gkai}“但是/然而”\end{CJK*} & 4 & 2 & 2 \\
& Causal Conjunctions & Cause-effect & ``because / therefore'' & \begin{CJK*}{UTF8}{gkai}“因为/因此”\end{CJK*} & 4 & 2 & 2 \\
& Conditional Conjunctions & Hypothetical conditions & ``if / unless'' & \begin{CJK*}{UTF8}{gkai}“如果/除非”\end{CJK*} & 4 & 2 & 3 \\
& Topicalizing Prepositions & Topic framing & ``regarding / about'' & \begin{CJK*}{UTF8}{gkai}“关于/对于”\end{CJK*} & 4 & 3 & 4 \\ \midrule
\multirow{2}{*}{\makecell[l]{7: Quantificational\\Details}}
& Specific Numerals & Exact numbers/dates & ``2023 / 70\%'' & \begin{CJK*}{UTF8}{gkai}“2023年/70\%”\end{CJK*} & 5 & 5 & 5 \\
& Vague Quantifiers & Approximate quantities & ``many / some'' & \begin{CJK*}{UTF8}{gkai}“很多/一些”\end{CJK*} & 3 & 4 & 4 \\ \midrule
\multirow{2}{*}{\makecell[l]{8: Punctuation\\Markers}}
& Neutral Punctuation & Standard delimiters & ``, / .'' & \begin{CJK*}{UTF8}{gkai}“，/。”\end{CJK*} & 5 & 5 & 5 \\
& Emphatic Punctuation & Emotional/exclamatory marks & ``! / …'' & \begin{CJK*}{UTF8}{gkai}“！/……”\end{CJK*} & 2 & 2 & 1 \\ \midrule
9: Modal Particles & Sentence-Final Particles & Attitudinal/pragmatic markers & ``ma / ne'' & \begin{CJK*}{UTF8}{gkai}“吗/呢”\end{CJK*} & 3 & 1 & 1 \\
\bottomrule
\end{tabularx}
\caption{Nine-layer linguistic hierarchy for discriminatory discourse analysis, with category/function, bilingual examples, and per-dimension importance levels (1 = high, 5 = low). These levels map to Table~\ref{tab:weight_scheme} and are not final sample ratings.}
\label{tab:linguistic_summary}
\end{table*}

\section{Methods}

\subsection{Overview of MAAM}

MAAM (Myopia--Astigmatism Anchor Mechanism) had two branches: Myopia for input-side anchor-preserving compression and Astigmatism for output-side contextual calibration (Appendix Figure~\ref{fig:maam_major}).
All three prediction dimensions shared this pipeline (Figure~\ref{fig:maam_pipeline}). We kept this design because it added no trainable parameters and left the backbone unchanged.
\subsection{Myopia}
\label{sec:myopia}

\begin{figure}[t]
\centering
\includegraphics[width=0.98\linewidth]{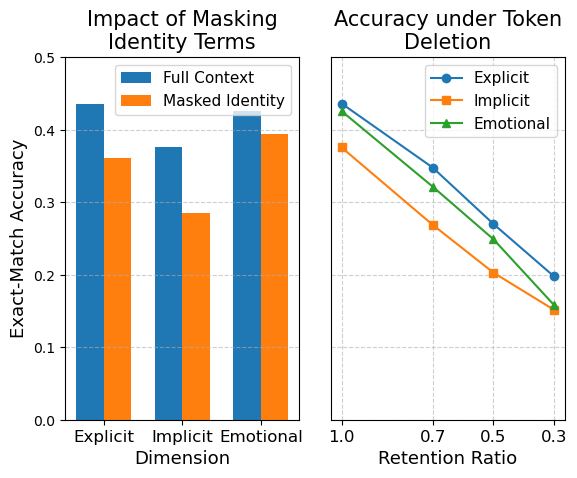}
\caption{Diagnostic evidence from preliminary experiments: random deletion and identity-anchor removal both hurt performance, with stronger degradation when anchor tokens are disrupted.}
\label{fig:qianzhi_anchor}
\end{figure}

Myopia performs anchor-preserving compression using a nine-layer linguistic taxonomy (Table~\ref{tab:linguistic_summary}) and its level-to-weight mapping (Table~\ref{tab:weight_scheme}).
Diagnostics (Figure~\ref{fig:qianzhi_anchor}; Appendix~\ref{app:anchor_diagnostic}) showed that random deletion hurt performance, with the largest drop when identity anchors were removed, confirming the need for anchor-preserving weighting.

The nine-layer taxonomy was motivated by linguistic and cognitive accounts of Chinese discriminatory discourse. It organizes cues by their roles in meaning construction: who is targeted, what attributes or stereotypes are attached, what stance or action is expressed, how affect is intensified, and how discourse context frames the utterance. The dimension-specific 1--5 levels in Table~\ref{tab:linguistic_summary} were then assigned through expert linguistic analysis of the ChLGBT task. Explicit bias prioritized overt identity arguments, evaluative modifiers, core predicates, and negation; implicit bias emphasized pragmatic or stereotype-bearing cues such as implicit arguments, contextual frames, and modal particles; emotional intensity prioritized affect-bearing modifiers, intensifiers, emphatic punctuation, and stance-bearing predicates. These levels provide a task-specific relevance map for bounded token weighting rather than universal constants, and can be recalibrated for other domains or languages by revising the subcategory-to-level mapping without changing the model architecture.

Operationally, POS/dependency parsing mapped each token to a linguistic subcategory, whose dimension-specific relevance level (1--5) was then converted into a bounded continuous weight.

\subsubsection{Keyword-Aware Weight Promotion}
Let $t_i$ denote the $i$-th token (or n-gram), and let $s_i \in [0,1]$ denote its association strength with a sensitive keyword set $\mathcal{K}$ (see Table~\ref{tab:top30_keywords} for representative entries).
Here, $i$ was the token index in a sentence, $N$ was sentence length, and $\mathcal{K}$ was the manually constructed sensitive lexicon used for keyword-aware promotion; we discuss ambiguous and coded identity terms in Appendix~\ref{app:coded_identity_terms}.

\paragraph{(1) Keyword Match Detection}
We set $s_i = 1$ if $t_i \in \mathcal{K}$ or there exists an $n$-gram $g = (t_i, \dots, t_{i+n-1})$ such that $g \in \mathcal{K}$ (and $s_i = 0$ otherwise).

\paragraph{(2) Sensitivity Level Promotion}
We defined a promotion flag $b_i = \mathbb{I}(s_i = 1)$. The final sensitivity level was
\begingroup
\setlength{\abovedisplayskip}{4pt}
\setlength{\belowdisplayskip}{4pt}
\small
\[
\ell_i^{\text{final}} =
\begin{cases}
\ell_{\min}, & \text{if } b_i = 1, \\
\ell_i^{\text{orig}}, & \text{otherwise},
\end{cases}
\]
\endgroup
where $\ell_{\min}$ was the most sensitive level and $\ell_i^{\text{orig}}$ was the parser-based level.
Levels were ordinal (1 highest, 5 lowest), so promotion to $\ell_{\min}$ assigned the highest semantic importance.

\paragraph{(3) Base Weight Assignment}
The base weight mapping satisfies $w_0(\ell_{\min}) > w_0(\ell)$ for all $\ell > \ell_{\min}$:
\[
w_i^{\text{base}} = w_0(\ell_i^{\text{final}}).
\]
Here, $w_0(\cdot)$ was the deterministic level-to-weight lookup defined by Table~\ref{tab:weight_scheme}, and $w_i^{\text{base}}$ was the pre-modulation weight for token $t_i$.

\paragraph{(4) Association-Aware Weight Modulation}

The final weight was modulated by the association strength $s_i$:

\begingroup
\setlength{\abovedisplayskip}{4pt}
\setlength{\belowdisplayskip}{4pt}
\small
\[
w_i^{\text{adj}} =
\begin{cases}
\min\big( w_i^{\text{base}} (1 + \beta s_i),\; u(\ell_i^{\text{final}}) \big), \\
\quad \text{if } s_i > \gamma, \\[2pt]
\max\big( w_i^{\text{base}} (1 - \varepsilon (1 - s_i)),\; l(\ell_i^{\text{final}}) \big), \\
\quad \text{otherwise},
\end{cases}
\]
\endgroup
where $\gamma$ was a threshold, $\beta$ and $\varepsilon$ controlled up/down adjustment, and $l(\cdot), u(\cdot)$ were level bounds.
Concretely, $\gamma$ decided whether a token entered the promotion branch ($s_i>\gamma$) or attenuation branch, $\beta$ was the promotion gain, $\varepsilon$ was the attenuation gain, and $l(\ell),u(\ell)$ were lower/upper allowable weights for level $\ell$. Hyperparameter selection and ablation protocol for these terms were provided in Appendix~\ref{app:myopia_hparam}. We used $\gamma=0.5$, $\beta=0.4$, and $\varepsilon=0.10$ for keyword-aware modulation. 

\paragraph{(5) Final Weight for Exact Keyword Matches}
For exact keyword matches ($s_i = 1$), the expression simplifies to
\begin{equation}
    w_i = \min\big( w_0(\ell_{\min}) (1 + \beta),\; u(\ell_{\min}) \big).
\end{equation}
Thus, exact sensitive matches were emphasized but remained bounded.
In this equation, $w_i$ was the final keyword-adjusted weight before smoothing for token $t_i$.

\subsubsection{Context-Aware Weight Smoothing}
We smoothed raw weights $\{w_i\}_{i=1}^N$ to suppress isolated spikes while preserving clustered cues.

\paragraph{(1) Local High-Weight Density Estimation}
Using a sliding window of size $k$, we computed high-weight density
\[
d_j = \frac{1}{k} \sum_{i=j}^{j+k-1} \mathbb{I}(w_i \geq \tau_h), \quad j = 1, \dots, N-k+1,
\]
where $\tau_h$ was a high-weight threshold.
Here, $k$ was the sliding-window length, $j$ indexed window start position, and $d_j$ measured the local proportion of salient tokens in window $j$.

\paragraph{(2) Adaptive Gating Coefficient}

Let $\bar{d}$ be the mean density and $d_{\max}=\max_j d_j$. Define dispersion $\Delta_D=d_{\max}-\bar{d}$. We computed

\begingroup
\setlength{\abovedisplayskip}{4pt}
\setlength{\belowdisplayskip}{4pt}
\small
\begingroup
\setlength{\abovedisplayskip}{4pt}
\setlength{\belowdisplayskip}{4pt}
\small
\[
\begin{aligned}
z &= \frac{1}{1 + \exp\big(\kappa (\Delta_D - \tau_f)\big)}, \\
\alpha &= z\,(\alpha_{\max}-\alpha_{\min}) + \alpha_{\min},
\end{aligned}
\]
\endgroup
\endgroup
where $\tau_f$ was a dispersion threshold and $\kappa$ controlled transition sharpness.
In addition, $z\in(0,1)$ was a logistic gate derived from dispersion, and $\alpha\in[\alpha_{\min},\alpha_{\max}]$ was the adaptive smoothing coefficient. In our implementation, smoothing used $\kappa=10$, $\tau_h=0.8$, $\tau_f=0.3$, with $\alpha_{\min}=0.3$ and $\alpha_{\max}=0.9$ (thus $\alpha\in[0.3,0.9]$).

\paragraph{(3) Bounded Convex Smoothing}
Each weight was first blended with the global mean $\bar{w}=\frac{1}{N}\sum_{i=1}^{N}w_i$:
\[
\tilde{w}_i = \alpha w_i + (1-\alpha)\bar{w}.
\]
Here, $\bar{w}$ was the sentence-level mean token weight, and $\tilde{w}_i$ was the intermediate blended weight before bound clipping.
Then it was clipped to the admissible interval $[l_{\ell_i},u_{\ell_i}]$:
\begin{equation}
    w_i^{\text{smooth}} = \min\big(u_{\ell_i},\; \max(l_{\ell_i},\; \tilde{w}_i)\big).
\end{equation}
In this final step, $\ell_i$ was the (possibly promoted) level for token $t_i$, while $w_i^{\text{smooth}}$ was the value passed to subsequent weighted pooling.

This design down-weighted unsupported isolated peaks (low $\alpha$) and preserved clustered salient cues (high $\alpha$). Tokens with level 5 (weight 0) were removed before encoding; levels 1--4 were retained and weighted during sentence pooling. Importantly, the downstream predictor consumed the smoothed weights $w_i^{\text{smooth}}$ (after keyword-aware modulation and smoothing), not the pre-smoothed weights. 

\begin{figure*}[t]
    \centering
    \includegraphics[width=0.98\linewidth]{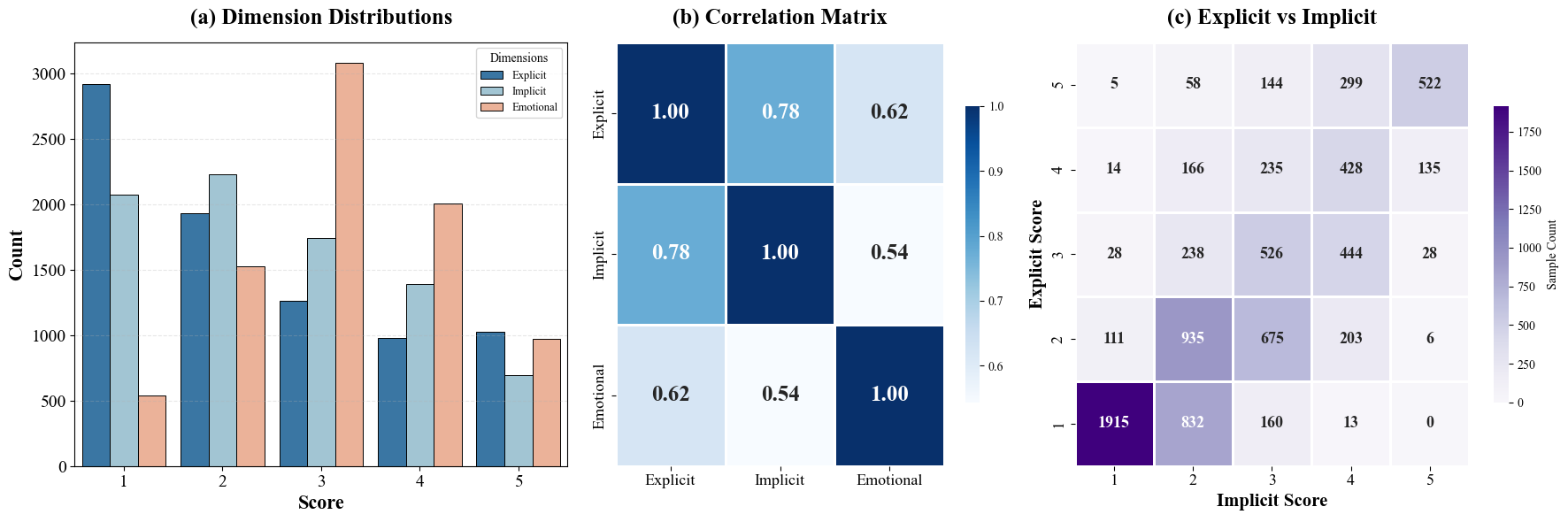}
    \caption{Main-text overview of ChLGBT, including label distribution, explicit--implicit coupling, and cross-dimension correlation.}
    \label{fig:dataset_overview_main}
\end{figure*}

\subsection{Astigmatism}
\label{sec:astigmatism}

Subjective judgments vary systematically with social context (e.g., tone, group identity, and stance). We modeled this effect with Astigmatism, a lightweight, model-agnostic module that encodes contextual bias as an explicit prior for calibration.

Unlike Myopia, which targeted surface-semantic anchoring, Astigmatism modeled context-induced shifts in judgment and fused them as computable priors with any downstream predictor.

\vspace{0.5em}
\paragraph{C--I--S Contextual Space}
We defined a three-axis contextual space reflecting common sources of judgment variation:
\begin{itemize}
    \item \textbf{Contextual Tone (C)}: expressive style (\texttt{Funny}, \texttt{General}, \texttt{Serious});
    \item \textbf{Group Identity (I)}: perceived affiliation (\texttt{Ingroup}, \texttt{Outgroup}, \texttt{Uncertain});
    \item \textbf{Stance Polarity (S)}: attitudinal orientation (\texttt{Pro}, \texttt{Anti}, \texttt{Neutral}).
\end{itemize}
Each dimension represented an independent source of systematic bias in human annotation.

\paragraph{Source of C–I–S Labels}
C--I--S labels were obtained either from external annotations or from text inference with NLP methods. In experiments, we used text inference to avoid additional manual metadata annotation. Method-specific details were provided in Section~\ref{sec:experiments} and Appendix~\ref{app:astigmatism}.

\vspace{0.5em}
\paragraph{Conditional Prior Estimation}

Given a labeled dataset $\mathcal{D} = \{(x_i, c_i, i_i, s_i, y_i)\}_{i=1}^N$, where $y_i \in \{1,\dots,5\}$ denoted an annotation label (explicit bias, implicit bias, or emotional intensity), we estimated empirical conditional distributions:

Here, $x_i$ was the $i$-th text sample, $(c_i,i_i,s_i)$ were its C--I--S labels, and $N$ was the number of labeled samples used to build priors.

\begingroup
\setlength{\abovedisplayskip}{4pt}
\setlength{\belowdisplayskip}{4pt}
\small
\begingroup
\setlength{\abovedisplayskip}{4pt}
\setlength{\belowdisplayskip}{4pt}
\small
\begin{equation}
    P(Y = k \mid Z = z) =
    \frac{\sum_{i=1}^N \mathbb{I}(y_i = k \land z_i = z)}
         {\sum_{i=1}^N \mathbb{I}(z_i = z)},
\end{equation}
\endgroup
\endgroup
where $Z \in \{C, I, S\}$. This yields three contextual prior libraries $\mathcal{Q}^C$, $\mathcal{Q}^I$, and $\mathcal{Q}^S$ (details in Appendix~\ref{app:astigmatism}).
In this formula, $Y$ was the target score variable, $k$ indexed one of the five score classes, $z$ was a concrete contextual state (e.g., \texttt{Serious}, \texttt{Outgroup}, \texttt{Anti}), and $z_i$ was the observed state for sample $i$ under the selected axis $Z$.
As an illustration of contextual prior calibration, the full C--I--S conditional score distributions are provided in Appendix Figure~\ref{fig:cis_heatmap}.

\vspace{0.5em}
\paragraph{Model-Agnostic Prior Fusion}
Let $\mathcal{M}$ be any probabilistic predictor with output $\mathbf{p}^{\mathcal{M}}(x)\in\mathbb{R}^5$. For contextual state $z(x)$, we retrieve the matched prior $\mathbf{q}^{(z)}\in\mathbb{R}^5$ and calibrate:
\begin{equation}
    \mathbf{p}^{\text{calib}}(x) = \alpha \mathbf{p}^{\mathcal{M}}(x) + (1 - \alpha) \mathbf{q}^{(z)},
\end{equation}
where $\alpha \in [0,1]$ was tuned on the validation set. We used convex fusion for robustness to noisy C/I/S labels and zero retraining cost (Appendix~\ref{app:astigmatism}). $\mathbf{p}^{\text{calib}}(x)$ was the final distribution used for downstream decision/scoring.

\subsection{Model Prediction and Integration}

This subsection clarifies the \textit{Model Prediction} and \textit{Integration} blocks in Figure~\ref{fig:maam_pipeline}.

\paragraph{Model Prediction}
The \textit{Model Prediction} block denoted the forward pass of a replaceable downstream model $\mathcal{M}$.
Any predictor that output a probability vector $\mathbf{p}^{\mathcal{M}}(x)$ could serve as the backbone.
MAAM did not rely on model-specific parameters or internal states, so replacing $\mathcal{M}$ did not require redesigning the framework.

\paragraph{Integration}
The \textit{Integration} block indicated how MAAM interfaced with the backbone: Myopia first filtered out level-5 tokens (weight $0$) before encoder input and then reweighted retained token contributions at aggregation time, while Astigmatism calibrated $\mathbf{p}^{\mathcal{M}}(x)$ with C--I--S priors after prediction.
Because both operations were external to model internals, the same integration pattern applied across backbones without architectural changes.

\section{Experiments}
\label{sec:experiments}

\begin{table*}[t]
\centering
\small
\setlength{\tabcolsep}{6pt}
\renewcommand{\arraystretch}{1.000}
\resizebox{\textwidth}{!}{%
\begin{tabular}{l l c c c c c c c c}
\toprule
\textbf{Backbone} & \textbf{Method / Variant} & \textbf{Exp. Acc} & \textbf{Exp. F1} & \textbf{Imp. Acc} & \textbf{Imp. F1} & \textbf{Emo. Acc} & \textbf{Emo. F1} & \textbf{BS} & \textbf{ECE} \\
\midrule
MacBERT & Baseline & 0.41$\pm$0.04 & 0.34$\pm$0.03 & 0.33$\pm$0.02 & 0.32$\pm$0.03 & 0.35$\pm$0.04 & 0.32$\pm$0.02 & 0.737 & 0.082 \\
 & w/o Myopia & 0.45$\pm$0.03 & 0.45$\pm$0.04 & 0.43$\pm$0.02 & 0.44$\pm$0.03 & 0.47$\pm$0.03 & 0.42$\pm$0.02 & 0.686 & 0.077 \\
 & w/o Astigmatism & 0.45$\pm$0.04 & 0.40$\pm$0.03 & 0.40$\pm$0.03 & 0.37$\pm$0.02 & 0.45$\pm$0.04 & 0.44$\pm$0.03 & 0.675 & 0.076 \\
 & MAAM (Full) & \textbf{0.48*$\pm$0.05} & 0.46*$\pm$0.05 & 0.44*$\pm$0.01 & 0.45*$\pm$0.02 & \textbf{0.52*$\pm$0.02} & 0.47*$\pm$0.01 & \textbf{0.622*} & 0.066* \\
\midrule
RoBERTa & Baseline & 0.39$\pm$0.03 & 0.39$\pm$0.04 & 0.32$\pm$0.02 & 0.29$\pm$0.03 & 0.36$\pm$0.03 & 0.36$\pm$0.02 & 0.752 & 0.091 \\
 & MAAM (Full) & \textbf{0.48*$\pm$0.02} & \textbf{0.47*$\pm$0.03} & \textbf{0.46*$\pm$0.02} & \textbf{0.46*$\pm$0.01} & 0.44*$\pm$0.03 & \textbf{0.48*$\pm$0.02} & \textbf{0.611*} & \textbf{0.059*} \\
\bottomrule
\end{tabular}
}
\caption{Merged comparison of ablation variants and backbone transfer with Acc, F1, Brier Score (BS), and corrected ECE. MacBERT rows use PT-MacBERT as the baseline checkpoint; ``w/o Myopia'' keeps only Astigmatism and ``w/o Astigmatism'' keeps only Myopia. Asterisks indicate statistically significant improvements over the corresponding baseline.}
\label{tab:ablation}
\label{tab:backbone}
\end{table*}

\begin{table*}[t]
\centering
\scalebox{0.73}{%
\begin{tabular}{lcccccc}
\toprule
\textbf{Model} & \textbf{Exp. Acc} & \textbf{Exp. F1} & \textbf{Imp. Acc} & \textbf{Imp. F1} & \textbf{Emo. Acc} & \textbf{Emo. F1} \\
\midrule
ChatGPT-4o & 0.47$\pm$0.07 & 0.42$\pm$0.08 & 0.42$\pm$0.02 & 0.42$\pm$0.03 & 0.42$\pm$0.02 & 0.44$\pm$0.02 \\
LLaMA-3.3 & 0.35$\pm$0.03 & 0.35$\pm$0.04 & 0.35$\pm$0.02 & 0.36$\pm$0.02 & 0.40$\pm$0.03 & 0.40$\pm$0.02 \\
ERNIE-5.0 & 0.40$\pm$0.05 & 0.35$\pm$0.07 & 0.41$\pm$0.01 & 0.41$\pm$0.01 & 0.47$\pm$0.02 & \textbf{\boldmath 0.47$\pm$0.01} \\
DeepSeek-V3.2 & 0.47$\pm$0.06 & 0.42$\pm$0.07 & 0.42$\pm$0.01 & 0.42$\pm$0.02 & 0.46$\pm$0.03 & 0.46$\pm$0.03 \\
Qwen3-Max & 0.45$\pm$0.06 & 0.39$\pm$0.07 & 0.38$\pm$0.03 & 0.37$\pm$0.02 & 0.40$\pm$0.03 & 0.40$\pm$0.03 \\
\midrule
MAAM (Ours) & \textbf{\boldmath 0.48$\pm$0.05} & \textbf{\boldmath 0.46$\pm$0.05} & \textbf{\boldmath 0.44*$\pm$0.01} & \textbf{\boldmath 0.45*$\pm$0.02} & \textbf{\boldmath 0.52*$\pm$0.02} & \textbf{\boldmath 0.47$\pm$0.01} \\
\bottomrule
\end{tabular}%
}
\caption{Zero-shot LLM performance on the ChLGBT test set, reported as mean$\pm$standard deviation over five independent runs. Asterisks indicate statistically significant improvements over the compared LLM baselines.}
\label{tab:llm_comparison}
\end{table*}

\subsection{Dataset}

Experiments were conducted on the ChLGBT dataset (Appendix~\ref{app:chlgbt_dataset}), which contains 8,120 annotated Chinese texts.
ChLGBT was the first fine-grained Chinese annotated dataset in this domain.
Most existing Chinese discriminatory-language datasets use binary or coarse-grained labels, limiting analysis of intensity, implicitness, and affective stance, especially for indirect or context-dependent prejudice.

Together with the scarcity of Chinese LGBT-focused resources, this motivates ChLGBT as a fine-grained, multi-dimensional dataset with an explicit scoring protocol.
Each sample was labeled on three dimensions using a 1--5 ordinal rating scale:
\begin{itemize}
    \item \textbf{Explicit bias} measures overt derogation, hostility, exclusion, or direct discriminatory expression toward LGBT individuals or groups.
    \item \textbf{Implicit bias} captures indirect prejudice conveyed through stereotypes, insinuation, pragmatic framing, or seemingly neutral statements with discriminatory implications.
    \item \textbf{Emotional intensity} measures the affective strength of the expression, regardless of whether the stance is discriminatory, supportive, ironic, or neutral.
\end{itemize}
For example, one ChLGBT post states: \begin{CJK*}{UTF8}{gkai}``一夫一妻制保证的是普通人的婚姻和生育权……在没有婚姻法保护下的同性恋，他能有好几个对象。''\end{CJK*} (``Monogamy protects ordinary people's marriage and reproductive rights; without marriage-law protection, a gay man could keep multiple partners.''). The post contains no overt slur (Explicit=2), but frames same-sex relationships through a stereotype of promiscuity and instability (Implicit=4). This low-explicit/high-implicit case illustrates why overt expression and pragmatically implied prejudice should be modeled separately. Additional contrastive examples are provided in Appendix Table~\ref{tab:explicit_implicit_contrast}.

Figure~\ref{fig:dataset_overview_main} presented the dataset overview used in this work.
We will release the non-identifying ChLGBT dataset, annotation guidelines, aggregate statistics, and access documentation at the project GitHub repository: \url{https://github.com/YuxinFu-NLP/MAAM-ChLGBT/tree/main}.
We used 1,000 samples for training and the rest for testing.

\subsection{Experimental Setup}

We used the pre-trained MacBERT~\cite{cui2020revisiting} checkpoint as the backbone and fine-tuned all supervised variants on the ChLGBT training split.
The baseline was fine-tuned on the original input, whereas Myopia variants used HanLP-derived token weights for input filtering and weighted pooling during supervised training.
For Astigmatism, C--I--S label inference was implemented with BGE embeddings (\texttt{bge-small-zh-v1.5} \cite{xiao2024c}; details in Appendix~\ref{app:astigmatism}).
Astigmatism was then applied after model prediction as posterior calibration, and all ablation variants were fine-tuned from the same checkpoint.

We reported Accuracy and Macro-F1 for task performance, and ECE/Brier score for calibration quality, with Macro-F1 as the primary metric.
Ablation, fusion-strategy, and zero-shot LLM results were shown in Tables~\ref{tab:ablation}, \ref{tab:best_fusion}, and \ref{tab:llm_comparison}, respectively.

\subsection{Myopia’s Impact on Representation}
\label{sec:myopia_analysis}

Myopia used parser-derived token weights $w_i \in [0,1]$ to form sentence representations: level-5 tokens were removed before encoding, and retained token states were aggregated by weighted pooling (Algorithm~\ref{alg:myopia}).
After filtering, retained character coverage was 88.57\% for explicit bias, 87.45\% for implicit bias, and 76.36\% for emotional intensity, indicating stronger pruning in the emotional dimension.

We further measured whether this weighting changed representations in the intended direction, with full metric definitions provided in Appendix~\ref{app:myopia_rep_metrics}.
On 8,120 sentences, MAD=0.401, indicating a substantial shift from mean pooling.
The weighted representation was also much closer to high-weight token subsets than to low-weight subsets ($0.0721 < 0.1754$, about $2.43\times$ closer), supporting that Myopia emphasizes salient tokens such as identity terms and stance-bearing predicates while suppressing less informative context.

\subsection{Fusion Strategy of Astigmatism}
\label{sec:fusion_strategy}

We treated C--I--S fusion order as a development-set selection problem rather than a manually fixed design choice. As a distributional reference, Appendix Figure~\ref{fig:cis_heatmap} presents the full C--I--S prior heatmaps used for calibration.

The full enumeration of single-label, pairwise, and triplet permutations is reported in Appendix~\ref{app:fusion_full} (Table~\ref{tab:full_fusion}), while Table~\ref{tab:best_fusion} summarizes the best development-set orders.

\begin{table}[t]
\centering
\small
\setlength{\tabcolsep}{5pt}
\renewcommand{\arraystretch}{1.0}
\begin{tabular}{lccc}
\toprule
\textbf{Dimension} & \textbf{Best Strategy} & \textbf{F1} & \textbf{$\Delta$ over Baseline} \\
\midrule
Explicit & I $\rightarrow$ S $\rightarrow$ C & 0.4623 & +0.0156 \\
Implicit & C $\rightarrow$ S $\rightarrow$ I & 0.4437 & +0.0062 \\
Emotional & I $\rightarrow$ S $\rightarrow$ C & 0.4687 & +0.0476 \\
\bottomrule
\end{tabular}
\caption{Performance of MAAM under different C--I--S fusion strategies.}
\label{tab:best_fusion}
\end{table}

\begin{table*}[t]
\centering
\scriptsize
\setlength{\tabcolsep}{4.2pt}
\renewcommand{\arraystretch}{0.95}
\resizebox{\textwidth}{!}{%
\begin{tabular}{lcccccccccc}
\toprule
\textbf{Field} & \textbf{\begin{CJK*}{UTF8}{gkai}月\end{CJK*}} & \textbf{\begin{CJK*}{UTF8}{gkai}半\end{CJK*}} & \textbf{\begin{CJK*}{UTF8}{gkai}，\end{CJK*}} & \textbf{\begin{CJK*}{UTF8}{gkai}骗\end{CJK*}} & \textbf{\begin{CJK*}{UTF8}{gkai}婚\end{CJK*}} & \textbf{\begin{CJK*}{UTF8}{gkai}同\end{CJK*}} & \textbf{\begin{CJK*}{UTF8}{gkai}性\end{CJK*}} & \textbf{\begin{CJK*}{UTF8}{gkai}恋\end{CJK*}} & \textbf{\begin{CJK*}{UTF8}{gkai}谈\end{CJK*}} & \textbf{\begin{CJK*}{UTF8}{gkai}！\end{CJK*}} \\
\midrule
\textbf{Gloss} & month & half & , & cheat & marriage & same & sex/gender & love/liaison & discuss & ! \\
\midrule
\textbf{BERT Baseline} & 0.036 & 0.015 & 0.170 & 0.013 & 0.013 & 0.020 & 0.014 & 0.021 & 0.012 & 0.367 \\
\textbf{MAAM Input ($w_i$)} & 0.000 & 0.000 & 0.000 & 0.879 & 0.879 & 0.879 & 0.879 & 0.879 & 0.817 & 0.629 \\
\textbf{Effective Weight ($\tilde{w}_i$)} & 0.000 & 0.000 & 0.000 & 0.086 & 0.086 & 0.086 & 0.086 & 0.086 & 0.080 & 0.062 \\
\textbf{Change ($\Delta$)} & \textbf{-0.036} & \textbf{-0.015} & \textbf{-0.170} & \textbf{+0.073} & \textbf{+0.074} & \textbf{+0.067} & \textbf{+0.072} & \textbf{+0.066} & \textbf{+0.069} & \textbf{-0.305} \\
\bottomrule
\end{tabular}%
}
\caption{Token-weight redistribution in the condensed case study for the sentence ``Yuepan (fatty), it's quite interesting to see a marriage-fraud gay talking about integrity!'' (\begin{CJK*}{UTF8}{gkai}“月半，骗婚的同性恋谈诚信挺有趣的！”\end{CJK*}). We report only tokens with the most salient positive/negative $\Delta$ shifts.}
\label{tab:weights_main}
\end{table*}

The best development-set order differs across tasks: Explicit and Emotional favor I$\rightarrow$S$\rightarrow$C, whereas Implicit favors C$\rightarrow$S$\rightarrow$I.
We interpret these differences as diagnostic evidence that contextual signals contribute differently across prediction dimensions, not as a claim of a universal causal ordering among C, I, and S.
For final test-set evaluation, we used a unified I$\rightarrow$S$\rightarrow$C strategy to avoid overfitting to dimension-specific development-set optima; Table~\ref{tab:ablation} reports the resulting test-set performance.

\subsection{Main Results}

\label{sec:main_results}

Table~\ref{tab:backbone} summarizes the ablation results and backbone-transfer performance on the ChLGBT test set.
On MacBERT, both single-module variants improved over the baseline, and the full MAAM configuration achieved the strongest overall performance across the three prediction dimensions.
The gains were accompanied by lower BS and ECE, suggesting improved calibration rather than only higher classification scores.
A similar improvement pattern was observed when replacing MacBERT with Chinese RoBERTa-wwm-ext~\cite{cui2020revisiting}, indicating that MAAM is not tailored to a single BERT-style backbone.

\paragraph{Comparison with Large Language Models and Backbone Robustness}
Table~\ref{tab:llm_comparison} compares five LLMs with the MacBERT-based MAAM configuration reported in Table~\ref{tab:ablation}, using the same fixed zero-shot protocol and prompt template shown in Table~\ref{tab:prompt}; all LLM scores are reported as mean$\pm$standard deviation over five runs.
MAAM obtained the highest explicit and implicit F1 scores, matched the best emotional F1 score, and achieved the highest accuracy on all three dimensions.
This comparison should be interpreted as a protocol-controlled reference rather than a claim that MAAM upper-bounds fully optimized LLM systems.
Additional few-shot LLM comparisons under 1-, 5-, and 10-shot settings are provided in Appendix~\ref{app:few_shot_llm}.

\paragraph{Cross-domain Generalization}
To test whether MAAM generalizes beyond ChLGBT, we additionally evaluated it on the COLD Region and Race subsets \cite{deng2022cold}.
MAAM consistently improved both task performance and calibration reliability; for example, on COLD Race, F1 increased from 0.906 to 0.928 while BS/ECE decreased from 0.074/0.076 to 0.057/0.054.
Full cross-domain results are reported in Appendix~\ref{app:cross_domain}.

\subsection{Condensed Case Study}
\label{sec:case_study_main}

We provide a compact in-text example to complement the full analysis in Appendix~\ref{sec:appendix_case_study}.
For the sentence ``Yuepan (fatty), it's quite interesting to see a marriage-fraud gay talking about integrity!''
(\begin{CJK*}{UTF8}{gkai}“月半，骗婚的同性恋谈诚信挺有趣的！”\end{CJK*}),
Table~\ref{tab:weights_main} shows that Myopia shifts weight from rhetorical or functional tokens toward discrimination-relevant anchors such as
\begin{CJK*}{UTF8}{gkai}骗婚\end{CJK*} and \begin{CJK*}{UTF8}{gkai}同性恋\end{CJK*}.
Astigmatism then calibrates the prediction under the inferred C--I--S context $(Funny, Outgroup, Anti)$.
Together, the case illustrates MAAM's complementarity: Myopia strengthens local discriminatory cues, while Astigmatism stabilizes final scoring through context-aware calibration.

\section{Conclusion}
We asked whether Chinese discriminatory-language detection could be improved without heavyweight scaling. Under our evaluation protocol, MAAM answers this question positively by combining Myopia-based anchor-preserving compression with Astigmatism-based C--I--S contextual calibration, yielding consistent gains over encoder baselines while remaining lightweight and interpretable.

On ChLGBT, MAAM improves accuracy and Macro-F1 while reducing Brier Score and ECE, suggesting better calibration as well as stronger classification. The gains transfer from MacBERT to Chinese RoBERTa-wwm-ext and extend to COLD Region and Race subsets, indicating applicability beyond the LGBT-focused setting. Compared with frontier LLMs, MAAM remains competitive under zero-shot and few-shot protocols, though these comparisons are protocol-bounded rather than claims against fully optimized LLM pipelines. Future work will test stronger backbones, broader bias domains, and finer-grained adaptive fusion.

\section*{Limitations}

This work has several limitations. First, although we evaluate MAAM on both ChLGBT and COLD subsets, broader validation across additional platforms, time periods, and bias categories is needed. Second, fine-grained discrimination labels, especially implicit bias and emotional intensity, inevitably involve contextual and cultural judgment despite multi-annotator adjudication. Third, MAAM relies on parser-derived token weights and corpus-estimated contextual priors, which may require recalibration when transferred to substantially different domains. Finally, our LLM comparison follows fixed zero-shot and few-shot protocols rather than exhaustive prompt or model-specific optimization, so the results should be interpreted within this evaluation setting.

\section*{Ethics Statement}
This work studies discriminatory language detection and includes potentially harmful or offensive content. We minimize risk by restricting data use to research purposes, removing direct user identifiers, excluding profile metadata and source URLs, and not releasing the original raw social-media texts. The non-identifying ChLGBT dataset will be released under the CC BY-NC-SA 4.0 license. Annotation was conducted by trained annotators with clear guidelines, and disagreements were resolved through adjudication to reduce arbitrary judgments. Because bias labels can reflect cultural and contextual subjectivity, we report labeling procedures and agreement statistics for transparency and emphasize that model outputs should support, not replace, human judgment in high-stakes settings. Our framework can be adapted across domains, but deployment should include domain-specific audits, periodic recalibration, and ongoing monitoring to mitigate unintended bias amplification.

\section*{Acknowledgments}
This work was supported by the Humanities and Social Sciences Research Youth Foundation of Ministry of Education of China under Grant Number 24YJCZH252, the 2024 Shanghai Educational Science Research Project (Special Project for Philosophy and Social Sciences Research in Shanghai Higher Education Institutions) entitled “Digital Technology Empowering the Great Founding Spirit of the Communist Party of China” (Grant No. 2024ZSW009), and the Undergraduate Research Mentorship Program at Shanghai International Studies University (Grant No. 41005391).

\bibliography{custom}

\appendix

\section{ChLGBT Dataset}
\label{app:chlgbt_dataset}

Existing Chinese discriminatory-language benchmarks are still dominated by binary 0/1 labels, which are useful for coarse detection but insufficient for analyzing the graded and multi-faceted nature of discriminatory discourse.
In particular, binary labels cannot separate overt hostility from subtle stereotyping, nor can they indicate whether a statement is emotionally intense, ironic, or pragmatically biased.
This limitation is especially consequential for LGBT-related Chinese discourse, where publicly available resources are scarce and harmful meanings are often implicit, contextual, or expressed through stereotypes rather than direct slurs.
Under these two constraints---the lack of fine-grained Chinese discrimination datasets and the scarcity of Chinese LGBT-focused resources---we constructed ChLGBT as a fine-grained, multi-dimensional dataset.
To support reproducibility and future comparison, we also designed detailed 1--5 scoring guidelines for explicit bias, implicit bias, and emotional intensity, which are summarized in Table~\ref{tab:annotation_guidance}.

\begin{table*}[ht]
\centering
\small
\renewcommand{\arraystretch}{1.2}
\begin{tabularx}{\textwidth}{|p{2cm}|p{0.5cm}|X|X|}
\hline
\textbf{Dimension} & \textbf{Score} & \textbf{Explanation} & \textbf{Example (English translation)} \\
\hline
Explicit & 1 & Mentions LGBT without hostility. & \begin{CJK*}{UTF8}{gkai}支持同性恋自由\end{CJK*} (Support LGBTQ+ freedom) \\
\hline
Explicit & 2 & Mild disrespect or problematic framing. & \begin{CJK*}{UTF8}{gkai}原来出生率低是因为同性恋率太高\end{CJK*} (Low birth rates are caused by high rates of homosexuality) \\
\hline
Explicit & 3 & Clearly unfriendly attitude toward LGBT. & \begin{CJK*}{UTF8}{gkai}长不成这样别说自己是同性恋\end{CJK*} (If you don't look like this, don't call yourself gay) \\
\hline
Explicit & 4 & Strong discriminatory remarks. & \begin{CJK*}{UTF8}{gkai}死小孩再鬼叫你长大变成同性恋\end{CJK*} (If you keep acting like a brat, you'll grow up to be gay!) \\
\hline
Explicit & 5 & Hostile, offensive, hateful language. & \begin{CJK*}{UTF8}{gkai}祝所有重男轻女的家庭生出来的男宝是同性恋得艾滋病\end{CJK*} (May all son-preference families have male heirs who turn out to be gay and contract AIDS) \\
\hline
Implicit & 1 & No Implicit; no indirect negativity. & \begin{CJK*}{UTF8}{gkai}我是同性恋爱好者\end{CJK*} (I'm a fan of gay people) \\
\hline
Implicit & 2 & Slight hints of bias, not obvious. & \begin{CJK*}{UTF8}{gkai}笑死我了真的受不了你们同性恋了\end{CJK*} (LMAO I really can't stand you gays) \\
\hline
Implicit & 3 & Clear Implicit via tone or context. & \begin{CJK*}{UTF8}{gkai}喜欢美国队长的男的都是同性恋啊\end{CJK*} (Guys who like Captain America are all gay) \\
\hline
Implicit & 4 & Strong Implicit using stereotypes. & \begin{CJK*}{UTF8}{gkai}最近真是邋遢我长得越来越像同性恋了\end{CJK*} (Lately I’m such a mess, I look more and more like a gay man) \\
\hline
Implicit & 5 & Deep Implicit and subtle discriminatory tone. & \begin{CJK*}{UTF8}{gkai}知不知道同性恋是精神疾病要治的！\end{CJK*} (Don’t you know homosexuality is a mental illness that needs treatment!) \\
\hline
Emotional & 1 & Calm and objective tone. & \begin{CJK*}{UTF8}{gkai}这对同性恋。\end{CJK*} (This is about gay people.) \\
\hline
Emotional & 2 & Slight Emotional lean, restrained. & \begin{CJK*}{UTF8}{gkai}知道你们是同性恋了\end{CJK*} (I know you're gay.) \\
\hline
Emotional & 3 & Clear Emotional stance. & \begin{CJK*}{UTF8}{gkai}这世界上真的有幸福的同性恋吗\end{CJK*} (Are there really happy gay couples in this world?) \\
\hline
Emotional & 4 & Strong Emotional tone. & \begin{CJK*}{UTF8}{gkai}同性恋真吓人!\end{CJK*} (Gay people are really scary!) \\
\hline
Emotional & 5 & Extremely strong emotions. & \begin{CJK*}{UTF8}{gkai}我他妈真想一拳打死同性恋！\end{CJK*} (I seriously want to punch a gay person to death!) \\
\hline
\end{tabularx}
\caption{Annotation scoring guidance for the three dimensions (1--5). \textbf{Note:} To ensure comparability and reproducibility, we use the same subtopic across different intensity levels; this does not imply the dataset is limited to homosexuality.}
\label{tab:annotation_guidance}

\end{table*}

\begin{table*}[t]
\centering
\scriptsize
\setlength{\tabcolsep}{3.5pt}
\renewcommand{\arraystretch}{1.15}
\resizebox{\textwidth}{!}{%
\begin{tabular}{p{2.3cm} p{7.0cm} c c p{6.2cm}}
\toprule
\textbf{Case Type} & \textbf{Original Text} & \textbf{Exp.} & \textbf{Imp.} & \textbf{Annotation Rationale} \\
\midrule
Low Exp. / High Imp. & \begin{CJK*}{UTF8}{gkai}越发理解一夫一妻制保证的是普通人的婚姻和生育权……在没有婚姻法保护下的同性恋，我甚至看到有些颜值和金钱加持下的男人，他能有好几个对象。\end{CJK*} (I now understand how monogamy protects ordinary people's marriage and reproductive rights; without marriage-law protection, I have even seen attractive or wealthy gay men keeping multiple partners.) & 2 & 4 & No overt slur or direct derogation, but the post frames same-sex relationships through a stereotype of promiscuity and instability. \\
High Exp. / Low Imp. & \begin{CJK*}{UTF8}{gkai}这天出门打球同性恋都能被冻直。\end{CJK*} (On a day this cold, even gay people would be frozen straight if they went out to play basketball.) & 4 & 2 & Uses overt identity-related hyperbole involving ``gay'' and ``turn straight,'' but the pragmatic target is exaggerating cold weather rather than developing a stereotype-based claim about LGBT people. \\
High Exp. / High Imp. & \begin{CJK*}{UTF8}{gkai}为什么同性恋这么多？是因为男人太低质，女人不想嫁，男人娶不着。物以类聚，人以群分。相似的人聚在一起，开始同性间的恋爱。\end{CJK*} (Why are there so many homosexuals? Because men are too low-quality, women do not want to marry them, and men cannot find wives. Birds of a feather flock together; similar people gather and begin same-sex relationships.) & 4 & 4 & Directly derogates homosexuality and explains it as the result of low-quality men failing in heterosexual marriage markets, combining overt contempt with a stigmatizing causal stereotype. \\
Low Exp. / Low Imp. & \begin{CJK*}{UTF8}{gkai}同性恋行为是由他的际遇决定，与他本人的意愿无关。既然与他本人无关，就不存在道德和思想问题。人的存在是一种自然现象，不是某种意志的产物。自然现象用美丑来划分更显得不郑重了。\end{CJK*} (Homosexual behavior is determined by one's circumstances and has nothing to do with personal will. Since it is unrelated to personal will, there is no moral or ideological problem. Human existence is a natural phenomenon, not the product of some will; judging a natural phenomenon as beautiful or ugly is even less appropriate.) & 1 & 1 & Mentions homosexuality in a de-moralizing and explanatory way; contains no overt derogation and explicitly rejects moral or ideological blame. \\
\bottomrule
\end{tabular}%
}
\caption{Contrastive Explicit--Implicit annotation examples. These cases show why the two labels are decoupled: Explicit captures surface-level identity-targeted expression, whereas Implicit captures stereotype-based or pragmatically implied prejudice.}
\label{tab:explicit_implicit_contrast}
\end{table*}

\subsection{Data Collection}
We collected Chinese-language posts from Weibo, one of the largest social media platforms in China, covering the period from January 1 to December 31, 2024. The data was collected using a publicly available Weibo scraping tool\footnote{\url{https://github.com/dataabc/weibo-search/}}. We used keyword-based queries related to LGBT topics, including but not limited to: “\begin{CJK*}{UTF8}{gkai}同性恋\end{CJK*}” (homosexual), “\begin{CJK*}{UTF8}{gkai}双性恋\end{CJK*}” (bisexual), “\begin{CJK*}{UTF8}{gkai}跨性别\end{CJK*}” (transgender), “\begin{CJK*}{UTF8}{gkai}非二元\end{CJK*}” (non-binary), “\begin{CJK*}{UTF8}{gkai}性少数\end{CJK*}” (sexual minority), “LGBT”, and “\begin{CJK*}{UTF8}{gkai}酷儿\end{CJK*}” (queer). Our keyword selection was designed to maximize retrieval precision while maintaining manageable data volume for manual annotation. Terms such as “\begin{CJK*}{UTF8}{gkai}同志\end{CJK*}” (comrade, colloquially used for gay individuals) were excluded due to high ambiguity in the Chinese Weibo context, where it could also refer to political comrades. The goal was to retrieve posts that either explicitly mentioned LGBT-related terms or were strongly associated with LGBT discourse for further manual filtering.

Only original posts were retained—comments, replies, and retweets were excluded to ensure linguistic independence and avoid duplicated context. This process resulted in a corpus of 8,120 unique posts for annotation.

\subsection{Data Cleaning}
To ensure the quality and relevance of our dataset, we conducted extensive data
cleaning on the collected posts. The cleaning process involved: (1) Deleting irrelevant
information, such as user details, creation timestamps, release locations, and device
usage; (2) Removing all forwarded posts to retain only the original microblogs; (3)
Stripping out emoticons and platform-specific content tags (e.g., ”@” mentions and
emojis); (4) Eliminating repetitive sentences to avoid redundancy; (5) Removing content
related to users’ personal privacy to protect individual identities; (6) Filtering
out texts in languages other than Chinese and English to maintain consistency.

We removed emojis and platform-specific symbols to reduce platform-dependent noise and keep the benchmark text-focused. This preprocessing rule was applied uniformly to all models and baselines. We acknowledged that this might discard some affective signal for emotional intensity, which we left to emoji-aware variants in future work.

\subsection{Manual Labeling}
We manually annotated the dataset with six native Chinese speakers (two LGBT members), all trained and blind to source metadata. Each annotator scored samples independently, and disagreements were resolved via discussion; final labels are rounded averages (to the nearest integer).

\paragraph{Inter-Annotator Agreement}
Inter-annotator reliability (Krippendorff's $\alpha$) was 0.73 (explicit), 0.67 (implicit), and 0.71 (emotional), indicating acceptable agreement. \citep{krippendorff2011computing}

\subsection{Annotation Protocol}
The main text defines the three annotation dimensions used in ChLGBT. Here we provide the operational annotation protocol: annotators scored each dimension independently on a 1--5 ordinal scale, using task-specific guidance to distinguish direct hostility, implied prejudice, and affective intensity. Table~\ref{tab:annotation_guidance} summarizes the score-level criteria and representative examples used during annotation.

\paragraph{Label Interpretation}
The 1--5 labels in ChLGBT should be interpreted as ordinal research labels assigned under the annotation protocol of this study, rather than objective or absolute measurements of discriminatory intent, implicit bias, or emotional intensity. Because discriminatory language can be context-dependent, implicit, ironic, or ambiguous, individual examples may admit alternative interpretations. Therefore, the labels are intended for dataset-level analysis, model training, and benchmark evaluation under the documented annotation scheme, and should not be treated as definitive moral, legal, or social judgments about any individual speaker or community.

\subsection{Label Statistics \& Analysis}
We summarized label distributions and inter-dimension relations for the three annotation axes (Figure~\ref{fig:dataset_overview_main}). Explicit bias and implicit bias were skewed toward lower scores, while emotional intensity was more evenly distributed. Correlations were positive but not redundant, indicating complementary signals across dimensions.

\subsection{Ethics Statement}
We remove all user identifiers (user IDs, handles, profile metadata) and do not release source URLs or original raw social-media texts. The released dataset contains non-identifying text content. Some posts may mention public figures; we treat these as publicly available references and do not add or infer any private information. We also remove personally identifying information where detected, but acknowledge that residual identifiers may remain in user‑generated text; users should not attempt re‑identification.

\section{ Mathematical Formulation and Diagnostics}

\subsection{Representation Metrics for Myopia}
\label{app:myopia_rep_metrics}

We used the same representation-diagnostic pipeline across dimensions and reported the Explicit dimension as a representative example in the main text. For each sentence, let $\mathbf{v}_{\mathrm{mean}}$ and $\mathbf{v}_{\mathrm{wgt}}$ denote mean-pooled and Myopia-weighted sentence vectors. We measured representation shift with mean absolute difference (MAD):
\[
\begin{aligned}
\mathrm{MAD} &= \frac{1}{768}\sum_{d=1}^{768}\left|v^{(d)}_{\mathrm{wgt}}-v^{(d)}_{\mathrm{mean}}\right|.
\end{aligned}
\]
Larger MAD indicates a stronger shift from mean pooling.

To test whether weighted representations align more closely with salient tokens, we construct $\mathbf{v}_{\mathrm{high}}$ by averaging only tokens with weights $>0.7$, and $\mathbf{v}_{\mathrm{low}}$ by averaging only tokens with weights $<0.5$. We then compute:
\[
\begin{aligned}
D_{\mathrm{high\text{-}wgt}} &= \frac{1}{768}\left\|\mathbf{v}_{\mathrm{high}}-\mathbf{v}_{\mathrm{wgt}}\right\|_{1}, \\
D_{\mathrm{low\text{-}wgt}}  &= \frac{1}{768}\left\|\mathbf{v}_{\mathrm{low}}-\mathbf{v}_{\mathrm{wgt}}\right\|_{1}.
\end{aligned}
\]
Here, $\|\cdot\|_1$ is the L1 norm. The intended pattern is $D_{\mathrm{high\text{-}wgt}} < D_{\mathrm{low\text{-}wgt}}$, meaning the final weighted representation is closer to high-weight semantic anchors than to low-weight contextual tokens.

\subsection{Top-30 LGBT-Related Keywords for Myopia Anchor Matching}
\label{app:top30_keywords}
To make keyword-based anchor matching explicit in the Myopia module, we list the top-30 LGBT-related keywords used in our retrieval and anchor construction in Table~\ref{tab:top30_keywords}.

\begin{table*}[t]
\centering
\small
\setlength{\tabcolsep}{5pt}
\renewcommand{\arraystretch}{1.12}
\begin{tabularx}{\textwidth}{@{} X X X X @{} }
\toprule
\textbf{Chinese Term} & \textbf{English Equivalent} & \textbf{Chinese Term} & \textbf{English Equivalent} \\
\midrule
1. \begin{CJK*}{UTF8}{gkai}同性恋\end{CJK*} & Homosexual / Gay \& Lesbian & 16. \begin{CJK*}{UTF8}{gkai}选定家庭\end{CJK*} & Chosen Family \\
2. \begin{CJK*}{UTF8}{gkai}双性恋\end{CJK*} & Bisexual & 17. \begin{CJK*}{UTF8}{gkai}去病理化\end{CJK*} & Depathologization \\
3. \begin{CJK*}{UTF8}{gkai}跨性别\end{CJK*} & Transgender & 18. \begin{CJK*}{UTF8}{gkai}扭转治疗\end{CJK*} & Conversion Therapy \\
4. \begin{CJK*}{UTF8}{gkai}酷儿\end{CJK*} & Queer & 19. \begin{CJK*}{UTF8}{gkai}骄傲月/游行\end{CJK*} & Pride Month / Parade \\
5. \begin{CJK*}{UTF8}{gkai}无性恋\end{CJK*} & Asexual & 20. \begin{CJK*}{UTF8}{gkai}彩虹旗\end{CJK*} & Rainbow Flag \\
6. \begin{CJK*}{UTF8}{gkai}泛性恋\end{CJK*} & Pansexual & 21. \begin{CJK*}{UTF8}{gkai}跨儿韧性\end{CJK*} & Trans Resilience \\
7. \begin{CJK*}{UTF8}{gkai}非二元性别\end{CJK*} & Non-binary & 22. \begin{CJK*}{UTF8}{gkai}性别代词\end{CJK*} & Pronouns \\
8. \begin{CJK*}{UTF8}{gkai}出柜\end{CJK*} & Coming out & 23. \begin{CJK*}{UTF8}{gkai}药娘\end{CJK*} & Transfeminine / GAHT users \\
9. \begin{CJK*}{UTF8}{gkai}深柜\end{CJK*} & In the closet & 24. \begin{CJK*}{UTF8}{gkai}TXL / 通讯录\end{CJK*} & Community Shorthand / TXL \\
10. \begin{CJK*}{UTF8}{gkai}性别焦虑\end{CJK*} & Gender dysphoria & 25. \begin{CJK*}{UTF8}{gkai}性别肯定手术\end{CJK*} & Gender Affirming Surgery (GAS) \\
11. \begin{CJK*}{UTF8}{gkai}性别欣悦\end{CJK*} & Gender euphoria & 26. \begin{CJK*}{UTF8}{gkai}交叉性\end{CJK*} & Intersectionality \\
12. \begin{CJK*}{UTF8}{gkai}顺性别\end{CJK*} & Cisgender & 27. \begin{CJK*}{UTF8}{gkai}彩虹资本主义\end{CJK*} & Rainbow Capitalism \\
13. \begin{CJK*}{UTF8}{gkai}异性恋霸权\end{CJK*} & Heteronormativity & 28. \begin{CJK*}{UTF8}{gkai}柏拉图式取向\end{CJK*} & Queerplatonic \\
14. \begin{CJK*}{UTF8}{gkai}同性婚姻合法化\end{CJK*} & Same-sex Marriage Equality & 29. \begin{CJK*}{UTF8}{gkai}身份政治\end{CJK*} & Identity Politics \\
15. \begin{CJK*}{UTF8}{gkai}意定监护\end{CJK*} & Legal Guardianship & 30. \begin{CJK*}{UTF8}{gkai}自我认同\end{CJK*} & Self-identification \\
\bottomrule
\end{tabularx}
\caption{Top-30 LGBT-related keywords used for retrieval and Myopia anchor matching (two side-by-side Chinese--English columns for compact layout).}
\label{tab:top30_keywords}
\end{table*}

\subsection{Robustness to Ambiguous and Coded Identity Terms}
\label{app:coded_identity_terms}
Chinese LGBT-related discourse often contains ambiguous or coded identity expressions, such as \begin{CJK*}{UTF8}{gkai}“同志”\end{CJK*} (literally ``comrade'') or homophonic/community substitutions such as \begin{CJK*}{UTF8}{gkai}“通讯录”\end{CJK*} and \begin{CJK*}{UTF8}{gkai}“南通”\end{CJK*}. MAAM is not designed as a closed keyword-matching system. Instead, Myopia uses an expandable anchor set: domain experts can add emerging slang, homophones, and platform-specific coded terms to $\mathcal{K}$ without changing the model architecture. Table~\ref{tab:top30_keywords} is therefore an illustrative subset rather than an exhaustive vocabulary.

When a coded term is absent from $\mathcal{K}$, Myopia still retains robustness through contextual weighting. The smoothing and bounded-weight design prevent the representation from depending on a single isolated keyword; surrounding predicates, modifiers, stance cues, and affective markers can still receive nonzero weights. Finally, Astigmatism provides an output-side safety net by recalibrating predictions with contextual C--I--S priors. Thus, slang robustness comes from the combination of expandable anchors, bounded contextual weighting, and posterior calibration, rather than from exact lexical matching alone.

\subsection{Myopia Hyperparameter Ablation and Selection}
\label{app:myopia_hparam}
To make the Myopia design reproducible, we ran a dedicated ablation on the development split for keyword-aware modulation and smoothing parameters; the search space is summarized in Table~\ref{tab:myopia_hparam_space}. We used one-factor-at-a-time sweeps around a strong reference setting and reported macro F1 over the three dimensions (explicit/implicit/emotional). The final configuration was selected by macro F1, with lower run-to-run variance as a tie-breaker.

\begin{table}[t]
\centering
\small
\setlength{\tabcolsep}{4pt}
\renewcommand{\arraystretch}{1.08}
\begin{tabular}{@{} 
    >{\raggedright\arraybackslash}p{0.17\columnwidth}
    >{\raggedright\arraybackslash}p{0.25\columnwidth}
    >{\raggedright\arraybackslash}p{0.5\columnwidth}
@{}}
\toprule
\textbf{Parameter} & \textbf{Search Space} & \textbf{Selection Principle} \\
\midrule
$\gamma$ & \seqsplit{\{0.3, 0.4, 0.5, 0.6, 0.7\}} & Balance promotion recall vs. precision \\
$\beta$ & \seqsplit{\{0.1, 0.2,..., 1.0\}} & Maximize anchor-token gains without saturation \\
$\varepsilon$ & \seqsplit{\{0.05, 0.1, ..., 0.8\}} & Avoid over-penalizing weakly related context \\
$k$ & \seqsplit{\{3, 5, 7, 9\}} & Best local-density discrimination \\
$\tau_h$ & \seqsplit{\{0.4, 0.5, 0.6, 0.7, 0.8\}} & Stable high-weight density estimation \\
$\kappa$ & \seqsplit{\{2, 4, 6, 8, 10\}} & Smooth but responsive gating transition \\
$\tau_f$ & \seqsplit{\{0.05, 0.10, 0.15, 0.20, 0.30\}} & Robust dispersion thresholding \\
$\alpha_{\min}$ & \seqsplit{\{0.10, 0.20, 0.30\}} & Preserve minimum anchor signal \\
$\alpha_{\max}$ & \seqsplit{\{0.70, 0.80, 0.90\}} & Prevent over-smoothing of salient clusters \\
\bottomrule
\end{tabular}
\caption{Development-set ablation space for Myopia hyperparameters.}
\label{tab:myopia_hparam_space}
\end{table}

In practice, we first tuned modulation parameters $(\gamma,\beta,\varepsilon)$ with smoothing fixed, then tuned smoothing parameters $(k,\tau_h,\kappa,\tau_f,\alpha_{\min},\alpha_{\max})$ on top of the best modulation setting. This staged protocol reduced search cost and isolated each component's effect. In total, Myopia used nine hyperparameters; in the supplementary sensitivity analyses we explicitly visualized the three modulation parameters $(\gamma,\beta,\varepsilon)$ in Table~\ref{tab:myopia_sensitivity_unified}, while the six smoothing/fusion parameters were reported with their search ranges and final selected values. Final selected values were $\gamma=0.5$, $\beta=0.4$, $\varepsilon=0.10$, $k=5$, $\tau_h=0.8$, $\tau_f=0.3$, $\kappa=10$, $\alpha_{\min}=0.3$, and $\alpha_{\max}=0.9$.

\subsection{Algorithm}
\begin{algorithm}
\caption{Myopia: Weighted Sentence Representation}
\label{alg:myopia}
\begin{algorithmic}[1]
\Require Input text $X$, token weights $\{w_i\}_{i=1}^n$
\Ensure Sentence vector $\mathbf{h}_{\text{sent}}$
\State $I \gets \{i \mid w_i > 0\}$
\State $X' \gets [x_i]_{i \in I}$, $\ \mathbf{H} \gets \text{Encode}(X')$
\State $\tilde{w}_i \gets w_i / \sum_{j \in I} w_j$ for all $i \in I$
\State $\mathbf{h}_{\text{sent}} \gets \sum_{i \in I} \tilde{w}_i \cdot \mathbf{h}_i$
\State \Return $\mathbf{h}_{\text{sent}}$
\end{algorithmic}
\end{algorithm}
Here, $n$ was sequence length, $I$ was the retained token-index set after zero-weight filtering, $\mathbf{h}_i$ was the contextual embedding of token $i$, and $\tilde{w}_i$ was the normalized weight over $I$.

\begin{figure*}[t]
\centering
\includegraphics[width=\textwidth]{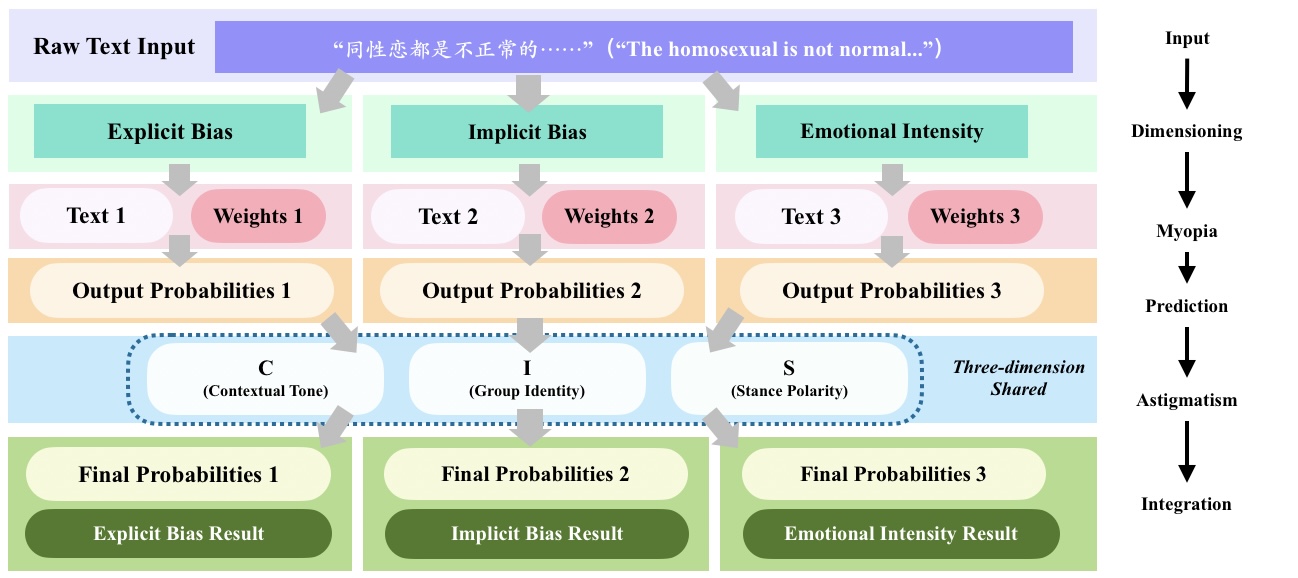}
\caption{MAAM Framework (macro-level overview). This figure highlights the parallel structure of MAAM, where the Myopia module performs input-side semantic compression and the Astigmatism module provides output-side contextual calibration, with both branches converging for final calibrated prediction.}
\label{fig:maam_major}
\end{figure*}

\subsection{Diagnostic Study on Topic-Sensitive Semantic Anchors}
\label{app:anchor_diagnostic}
To motivate MAAM’s design choice of preserving discriminative semantic anchors under compression, we ran two diagnostics on a high-context task: LGBT-related discrimination in Chinese social media. In this setting, identity terms (e.g., \begin{CJK*}{UTF8}{gkai}“跨性别”\end{CJK*}) acted as topic-sensitive semantic anchors that grounded the evaluative meaning of the whole utterance.

All experiments used the Qwen3-Max model on a random sample of 1,000 annotated posts drawn from Appendix~\ref{app:chlgbt_dataset} (no new data were collected). This pilot scale was large enough to expose stable patterns while keeping computation tractable.
Under our level-based Myopia filtering, the average retained character coverage (relative to the original text) was 88.57\% for explicit bias, 87.45\% for implicit bias, and 72.36\% for emotional intensity.

We tested two hypotheses:
\begin{enumerate}
    \item \textbf{Unstructured input reduction harms performance}, because random token deletion fails to distinguish between noise and anchor-bearing segments;
    \item \textbf{Masking topic-sensitive anchors degrades judgment}, even when surrounding context remains intact, revealing models’ reliance on these pivotal units for bias assessment.
\end{enumerate}

\noindent\textbf{Results for Hypothesis 1 (Random Token Deletion).}  
We randomly retained 100\%, 70\%, 50\%, and 30\% of tokens in each input and evaluated model accuracy. As shown in Table~\ref{tab:random_deletion}, performance dropped sharply as retention decreased—e.g., implicit bias accuracy fell from 0.3755 to 0.1515 when only 30\% of tokens remained. This confirmed that unstructured compression discarded discriminative signals, and models could not autonomously identify semantically critical tokens.

\begin{table}[h]
\centering

% 使用 tabularx，总宽度设为 \textwidth
% X 列会自动平分剩余宽度，并支持居中对齐（见下方说明）
\begin{tabularx}{\columnwidth}{@{}l *{3}{>{\centering\arraybackslash}X}@{}}
\toprule
Retention Ratio & Explicit & Implicit & Emotional \\
\midrule
1.0 & 0.4361 & 0.3755 & 0.4256 \\
0.7 & 0.3478 & 0.2688 & 0.3215 \\
0.5 & 0.2701 & 0.2029 & 0.2490 \\
0.3 & 0.1976 & 0.1515 & 0.1581 \\
\bottomrule
\end{tabularx}
\caption{Model exact-match accuracy under random token deletion at varying retention ratios (1,000 examples, Qwen3-Max).}
\label{tab:random_deletion}
\end{table}

\begin{table}[h]
\centering

\begin{tabularx}{\columnwidth}{@{}l *{3}{>{\centering\arraybackslash}X}@{}}
\toprule
Dimension & Full Context & Masked Identity & Drop \\
\midrule
Explicit & 0.4361 & 0.3610 & 0.0909 \\
Implicit & 0.3755 & 0.2846 & 0.0870 \\
Emotional & 0.4256 & 0.3939 & 0.0632 \\
\bottomrule
\end{tabularx}
\caption{Impact of masking topic-sensitive identity terms on model accuracy (1,000 examples, Qwen3-Max). Drop = Full – Masked.}
\label{tab:anchor_mask}
\end{table}

\noindent\textbf{Results for Hypothesis 2 (Anchor Masking).}  
We replaced identity-related terms (e.g., \begin{CJK*}{UTF8}{gkai}“同性恋”\end{CJK*}, \begin{CJK*}{UTF8}{gkai}“跨性别”\end{CJK*}) with a placeholder token \texttt{[IDENTITY]} while preserving sentence structure. As Table~\ref{tab:anchor_mask} showed, masking anchors reduced accuracy across all dimensions: 9.09\% (explicit bias), 8.70\% (implicit bias), and 6.32\% (emotional intensity). The drop—especially for implicit bias—showed that models relied on these lexical pivots even when surrounding context remained.

\noindent\textbf{Conclusion.}  
These findings validate both hypotheses and directly motivate MAAM’s core mechanisms: (1) structured compression that preserves syntactic and semantic skeletons (addressing Hypothesis 1), and (2) explicit anchoring of topic-sensitive units during blur simulation (addressing Hypothesis 2). Although this pilot focuses on LGBT discourse, the underlying challenge—preserving meaning-critical pivots under input perturbation—also appears in other bias domains; our later COLD Region and Race experiments further evaluate this cross-domain applicability. The full evaluation of MAAM on the complete 8,120 dataset is presented in Section~\ref{sec:main_results}.

\begin{table}[h]
\centering

\small
\begin{tabular}{ccc}
\toprule
Score Level & Assigned Weight & Retention Threshold \\
\midrule
1 & 0.90 & [0.8, 1.0] \\
2 & 0.60 & [0.5, 0.7] \\
3 & 0.30 & [0.2, 0.4] \\
4 & 0.05 & [0.0, 0.1] \\
5 & 0.00 & excluded pre-encoder \\
\bottomrule
\end{tabular}

\caption{Weight assignment scheme in MAAM's myopia simulation. A score level maps to an initial weight tier (e.g., level 2 maps to a 0.5--0.7 range with a default of 0.6). Subsequent smoothing can only adjust weights within the tier's range, unless Keyword-Aware Weight Promotion elevates the token to a higher tier.}
\label{tab:weight_scheme}
\end{table}

\begin{table}
\centering
\scriptsize
\setlength{\tabcolsep}{3.2pt}
\renewcommand{\arraystretch}{0.95}
\setlength{\tabcolsep}{5pt}
\renewcommand{\arraystretch}{1.0}
\begin{tabular}{p{0.95\linewidth}}
\toprule
\textbf{Prompt} \\
\midrule

\begin{CJK*}{UTF8}{gkai}
prompt = "请对以下文本进行分析，并评分： \{text\}
评分标准：1. 显性歧视 (1-5)   - 1分：没有任何显性歧视行为，文本内容中没有攻击性或侮辱性的语言。   - 2-4分：逐渐增高的歧视行为。  - 5分：文本包含非常明显的歧视言辞，语言攻击性强，表现出对同性恋群体的敌意或贬低。  2. 隐性歧视 (1-5)   - 1分：没有隐性歧视成分，文本内容不包含任何间接的贬低、偏见或隐蔽的负面情绪。   - 2-4分：逐渐增高的歧视行为。   - 5分：文本隐含非常深刻的歧视成分，可能以微妙的方式表现出对同性恋群体的偏见或歧视。3. 情绪激烈度 (1-5)   - 1分：文本情绪冷静、理性，没有明显的情感倾向。即使是讨论同性恋话题，也保持了客观、平和的语气。   - 2-4分：逐渐增高的情绪激烈度，表示情感从冷静中立到偏向支持或反对的激烈程度。  - 5分：文本情感极为强烈，表现出过度的支持或强烈的反对，语气激进或偏激。请返回一个字典，格式如下：\{\{”显性歧视“: 显性歧视分数, ”隐性歧视“: 隐性歧视分数, ”情感强烈程度“: 情感强烈程度分数\}\}"
\end{CJK*} \\
\bottomrule
\end{tabular}
\caption{Zero-shot prompt used for all LLM evaluations.}
\label{tab:prompt}
\end{table}

\begin{table*}[t]
\centering
\small

\begin{tabular}{llll}
\toprule
\textbf{Dimension} & \textbf{Strategy} & \textbf{F1 ($\Delta$)} & \textbf{$\alpha$} \\
\midrule
\multicolumn{4}{l}{Explicit (Baseline F1 = 0.4467)} \\
& [C]               & 0.4530 (+0.0063) & [0.3] \\
& [I]               & 0.4520 (+0.0052) & [0.4] \\
& [S]               & 0.4579 (+0.0111) & [0.4] \\
& [C \textrightarrow I]       & 0.4553 (+0.0086) & [0.4, 0.7] \\
& [I \textrightarrow C]       & 0.4559 (+0.0092) & [0.4, 0.7] \\
& [C \textrightarrow S]       & 0.4598 (+0.0131) & [0.4, 0.8] \\
& [S \textrightarrow C]       & 0.4620 (+0.0152) & [0.5, 0.6] \\
& [I \textrightarrow S]       & 0.4620 (+0.0153) & [0.4, 0.8] \\
& [S \textrightarrow I]       & 0.4605 (+0.0137) & [0.6, 0.6] \\
& [C \textrightarrow I \textrightarrow S]   & 0.4622 (+0.0154) & [0.7, 0.8, 0.6] \\
& [C \textrightarrow S \textrightarrow I]   & 0.4623 (+0.0155) & [0.6, 0.6, 0.9] \\
& [I \textrightarrow C \textrightarrow S]   & 0.4623 (+0.0156) & [0.7, 0.8, 0.6] \\
& [I \textrightarrow S \textrightarrow C]   & 0.4623 (+0.0156) & [0.7, 0.6, 0.7] \\
& [S \textrightarrow C \textrightarrow I]   & 0.4620 (+0.0152) & [0.5, 0.6, 1.0] \\
& [S \textrightarrow I \textrightarrow C]   & 0.4622 (+0.0154) & [0.6, 0.6, 0.9] \\
\midrule
\multicolumn{4}{l}{Implicit (Baseline F1 = 0.4375)} \\
& [C]               & 0.4421 (+0.0046) & [0.5] \\
& [I]               & 0.4400 (+0.0025) & [0.4] \\
& [S]               & 0.4424 (+0.0048) & [0.6] \\
& [C \textrightarrow I]       & 0.4422 (+0.0046) & [0.5, 0.9] \\
& [I \textrightarrow C]       & 0.4427 (+0.0052) & [0.7, 0.6] \\
& [C \textrightarrow S]       & 0.4429 (+0.0054) & [0.5, 0.9] \\
& [S \textrightarrow C]       & 0.4432 (+0.0057) & [0.9, 0.5] \\
& [I \textrightarrow S]       & 0.4431 (+0.0055) & [0.4, 0.6] \\
& [S \textrightarrow I]       & 0.4432 (+0.0056) & [0.4, 0.6] \\
& [C \textrightarrow I \textrightarrow S]   & 0.4433 (+0.0058) & [0.9, 0.5, 0.6] \\
& [C \textrightarrow S \textrightarrow I]   & 0.4437 (+0.0062) & [0.8, 0.4, 0.7] \\
& [I \textrightarrow C \textrightarrow S]   & 0.4433 (+0.0057) & [0.7, 0.9, 0.4] \\
& [I \textrightarrow S \textrightarrow C]   & 0.4432 (+0.0057) & [1.0, 0.9, 0.5] \\
& [S \textrightarrow C \textrightarrow I]   & 0.4432 (+0.0057) & [0.9, 0.5, 1.0] \\
& [S \textrightarrow I \textrightarrow C]   & 0.4432 (+0.0057) & [0.9, 1.0, 0.5] \\
\midrule
\multicolumn{4}{l}{Emotional (Baseline F1 = 0.4211)} \\
& [C]               & 0.4482 (+0.0271) & [0.3] \\
& [I]               & 0.4511 (+0.0300) & [0.3] \\
& [S]               & 0.4546 (+0.0335) & [0.3] \\
& [C \textrightarrow I]       & 0.4511 (+0.0300) & [1.0, 0.3] \\
& [I \textrightarrow C]       & 0.4526 (+0.0315) & [0.4, 0.7] \\
& [C \textrightarrow S]       & 0.4675 (+0.0464) & [0.9, 0.3] \\
& [S \textrightarrow C]       & 0.4642 (+0.0430) & [0.3, 0.9] \\
& [I \textrightarrow S]       & 0.4687 (+0.0476) & [0.9, 0.3] \\
& [S \textrightarrow I]       & 0.4668 (+0.0456) & [0.3, 0.9] \\
& [C \textrightarrow I \textrightarrow S]   & 0.4687 (+0.0476) & [1.0, 0.9, 0.3] \\
& [C \textrightarrow S \textrightarrow I]   & 0.4675 (+0.0464) & [0.9, 0.3, 1.0] \\
& [I \textrightarrow C \textrightarrow S]   & 0.4687 (+0.0476) & [0.9, 1.0, 0.3] \\
& [I \textrightarrow S \textrightarrow C]   & 0.4687 (+0.0476) & [0.9, 0.3, 1.0] \\
& [S \textrightarrow C \textrightarrow I]   & 0.4668 (+0.0456) & [0.3, 1.0, 0.9] \\
& [S \textrightarrow I \textrightarrow C]   & 0.4668 (+0.0456) & [0.3, 0.9, 1.0] \\
\bottomrule
\end{tabular}
\caption{Full ablation of C--I--S fusion strategies across three bias dimensions. $\Delta$ denotes absolute improvement over the baseline. Fusion weights $\alpha$ are searched in steps of 0.1 for computational efficiency.}
\label{tab:full_fusion}
\end{table*}

\subsection{Astigmatism Implementation Details}
\label{app:astigmatism}

\paragraph{0. C--I--S Label Inference}
C--I--S labels were automatically inferred from raw text using an embedding-based nearest-prototype method with \texttt{BAAI/bge-small-zh-v1.5} \cite{xiao2024c}. For each dimension (C/I/S), we prepared short prototype descriptions for each candidate label, computed sentence and prototype embeddings, and assigned the label with maximum cosine similarity. Concretely, the Contextual Tone prototypes described \texttt{Funny} as humorous, joking, ironic, or playful expression; \texttt{Serious} as formal, solemn, argumentative, or factual expression; and \texttt{General} as neutral or ordinary expression. The Group Identity prototypes distinguished \texttt{Ingroup}, \texttt{Outgroup}, and \texttt{Uncertain} speaker positions, while the Stance Polarity prototypes distinguished \texttt{Pro}, \texttt{Anti}, and \texttt{Neutral} attitudes. A validation-set threshold was used to route ambiguous cases to default categories (\texttt{General} for C, \texttt{Uncertain} for I, and \texttt{Neutral} for S).

\paragraph{Robustness to Inference Noise}
The C--I--S labels were used as soft calibration signals rather than gold-standard supervision or hard decision rules. Therefore, moderate errors in automatic label inference cannot directly override the backbone prediction. The contextual priors were estimated as empirical score distributions from the training data, and different labels often produced relatively small posterior shifts. For example, as shown in Appendix Figure~\ref{fig:cis_heatmap}, the Explicit-track Score-1 prior probabilities for \texttt{Funny}, \texttt{General}, and \texttt{Serious} contexts were close (0.354, 0.332, and 0.373, respectively), indicating a narrow influence gap between neighboring contextual labels. Together with convex fusion, this design makes Astigmatism a conservative recalibration layer: an incorrect C--I--S label may slightly adjust the final distribution, but it is unlikely to catastrophically change the prediction by itself.

\paragraph{1. Prior Construction}
For each contextual dimension $Z \in \{C, I, S\}$ and judgment type (explicit bias/implicit bias/emotional intensity), we computed empirical probabilities over rating levels $k \in \{1,\dots,5\}$. If no samples existed for a context--rating pair $(z, k)$, we assigned a uniform prior of $0.2$ (i.e., $P(Y=k \mid Z=z) = 0.2$).

\paragraph{2. Calibration Protocol}
We followed the Astigmatism calibration defined in Section~\ref{sec:astigmatism}, and only report the implementation protocol here. We explored two $\alpha$-selection variants:
\begin{itemize}
    \item \textbf{Per-class $\alpha$}: $\boldsymbol{\alpha} = [\alpha_1, \dots, \alpha_5]$, where each $\alpha_k$ was searched independently from $\{0.1, 0.2, \dots, 1.0\}$ to maximize F1 for class $k$;
    \item \textbf{Unified $\alpha$}: $\boldsymbol{\alpha} = [\alpha, \dots, \alpha]$, with $\alpha$ selected to maximize macro F1.
\end{itemize}
In this appendix setting, $\boldsymbol{\alpha}\in\mathbb{R}^5$ was either a per-class mixing vector or a shared scalar replicated across classes. After fusion, probabilities were clipped to $[10^{-8}, \infty)$ and renormalized to sum to 1.

\paragraph{3. Evaluation Metric}
Class-specific F1 scores used \texttt{zero\_division=0} (scikit-learn default), and macro F1 was reported for overall performance.

\section{Supplementary Analyses and Case Studies}
\subsection{Cross-Domain Generalization on COLD}
\label{app:cross_domain}

To assess whether MAAM transfers beyond the ChLGBT benchmark, we evaluated the same methodology on the COLD Region and Race subsets \cite{deng2022cold}. Table~\ref{tab:cold_generalization_appendix} reports both task performance and calibration reliability. Across both subsets, MAAM improves Acc/F1 while reducing BS and ECE, indicating that the calibration benefit is not specific to LGBT-focused discourse. Since Region and Race are separate COLD subsets with different data distributions, the comparison is intended to show consistent within-subset gains rather than to compare their absolute difficulty.

\begin{table*}[t]
\centering
\small
\resizebox{\textwidth}{!}{%
\begin{tabular}{lcccccccc}
\toprule
\multirow{2}{*}{\textbf{Method}} & \multicolumn{4}{c}{\textbf{Region}} & \multicolumn{4}{c}{\textbf{Race}} \\
\cmidrule(lr){2-5} \cmidrule(lr){6-9}
 & \textbf{Acc} & \textbf{F1} & \textbf{BS} & \textbf{ECE} & \textbf{Acc} & \textbf{F1} & \textbf{BS} & \textbf{ECE} \\
\midrule
Baseline & 0.885$\pm$0.009 & 0.876$\pm$0.008 & 0.124 & 0.119 & 0.903$\pm$0.010 & 0.906$\pm$0.007 & 0.074 & 0.076 \\
w/o Myopia & 0.888$\pm$0.008 & 0.878$\pm$0.008 & 0.108 & 0.078 & 0.914$\pm$0.009 & 0.919$\pm$0.008 & 0.064 & 0.060 \\
w/o Astigmatism & 0.890$\pm$0.006 & 0.880$\pm$0.010 & 0.113 & 0.107 & 0.915$\pm$0.009 & 0.919$\pm$0.008 & 0.066 & 0.060 \\
MAAM (Full) & \textbf{0.891$\pm$0.007} & \textbf{0.881$\pm$0.010} & \textbf{0.091} & \textbf{0.071} & \textbf{0.925$\pm$0.010} & \textbf{0.928$\pm$0.008} & \textbf{0.057} & \textbf{0.054} \\
\bottomrule
\end{tabular}
}
\caption{Generalization test on COLD Region and Race subsets (binary tasks), including Brier Score (BS) and corrected ECE as reliability metrics. ``w/o Myopia'' keeps only Astigmatism, and ``w/o Astigmatism'' keeps only Myopia.}
\label{tab:cold_generalization_appendix}
\end{table*}

\subsection{Myopia Hyperparameter Sensitivity ($\gamma,\beta,\varepsilon$)}
\label{app:myopia_sensitivity}

This subsection reports the merged sensitivity results for the three Myopia modulation parameters (BERT backbone, Astigmatism disabled); the tuning protocol and selection rationale are described in Section~\ref{app:myopia_hparam}.

We used a one-factor-at-a-time protocol: when sweeping one modulation parameter (e.g., $\gamma$), the other two were fixed, and all remaining Myopia hyperparameters kept their final values from the main text. Concretely, unless otherwise specified, the fixed reference setting was $\gamma=0.5$, $\beta=0.4$, $\varepsilon=0.10$, $k=5$, $\tau_h=0.8$, $\tau_f=0.3$, $\kappa=10$, $\alpha_{\min}=0.3$, and $\alpha_{\max}=0.9$.

\begin{table*}[ht]
\centering
\small

\begin{tabular}{llcccccccc}
\toprule
Param. & Value & Exp. Acc & Exp. F1 & Imp. Acc & Imp. F1 & Emo. Acc & Emo. F1 & Avg. Acc & Avg. F1 \\
\midrule
\multirow{3}{*}{$\gamma$}
& 0.3 & 0.4619 & 0.4574 & 0.4446 & 0.4485 & 0.4812 & 0.4328 & 0.4626 & 0.4462 \\
& \textbf{0.5} & \textbf{0.4608} & \textbf{0.4567} & \textbf{0.4466} & \textbf{0.4508} & \textbf{0.4832} & \textbf{0.4340} & \textbf{0.4635} & \textbf{0.4472} \\
& 0.7 & 0.4608 & 0.4567 & 0.4466 & 0.4508 & 0.4832 & 0.4340 & 0.4635 & 0.4472 \\
\midrule
\multirow{4}{*}{$\beta$}
& 0.2 & 0.4608 & 0.4567 & 0.4466 & 0.4508 & 0.4832 & 0.4340 & 0.4635 & 0.4472 \\
& \textbf{0.4} & \textbf{0.4608} & \textbf{0.4567} & \textbf{0.4466} & \textbf{0.4508} & \textbf{0.4832} & \textbf{0.4340} & \textbf{0.4635} & \textbf{0.4472} \\
& 0.6 & 0.4608 & 0.4567 & 0.4466 & 0.4508 & 0.4832 & 0.4340 & 0.4635 & 0.4472 \\
& 0.8 & 0.4608 & 0.4567 & 0.4466 & 0.4508 & 0.4832 & 0.4340 & 0.4635 & 0.4472 \\
\midrule
\multirow{5}{*}{$\varepsilon$}
& 0.05 & 0.4608 & 0.4565 & 0.4456 & 0.4492 & 0.4832 & 0.4347 & 0.4632 & 0.4468 \\
& \textbf{0.10} & \textbf{0.4608} & \textbf{0.4565} & \textbf{0.4466} & \textbf{0.4508} & \textbf{0.4832} & \textbf{0.4347} & \textbf{0.4635} & \textbf{0.4473} \\
& 0.15 & 0.4608 & 0.4567 & 0.4466 & 0.4508 & 0.4832 & 0.4340 & 0.4635 & 0.4472 \\
& 0.20 & 0.4598 & 0.4549 & 0.4466 & 0.4508 & 0.4832 & 0.4340 & 0.4632 & 0.4466 \\
& 0.25 & 0.4598 & 0.4549 & 0.4466 & 0.4508 & 0.4832 & 0.4340 & 0.4632 & 0.4466 \\
\bottomrule
\end{tabular}
\caption{Unified sensitivity table for Myopia modulation hyperparameters under one-factor-at-a-time sweeps (one of $\gamma$, $\beta$, or $\varepsilon$ varies while the others are fixed), with all non-swept Myopia hyperparameters fixed to the final setting in the main text. Bold rows indicate the final selected values used in our main experiments ($\gamma=0.5$, $\beta=0.4$, $\varepsilon=0.10$). Results are reported on the development split only (not the full test set).}
\label{tab:myopia_sensitivity_unified}

\end{table*}

Overall, $\gamma$ showed a mild plateau effect ($0.5\approx0.7$), $\beta$ was effectively insensitive in the tested range, and $\varepsilon$ exhibited only minor variation with best average macro-F1 at $\varepsilon=0.10$.

\subsection{Full C--I--S Fusion Strategy Ablation}
\label{app:fusion_full}

We exhaustively evaluated all single, pairwise, and triple cascading fusion orders of the C (Contextual Tone), I (Group Identity), and S (Stance Polarity) contextual signals. For each strategy, we performed a grid search over fusion weights $\alpha$ with step size 0.1 (i.e., $\alpha \in \{0.1, 0.2, \dots, 1.0\}$) to balance the original model logits and contextual priors. This step size was a simplified and computationally efficient choice for ablation; more precise tuning could use finer-grained $\alpha$ values (e.g., 0.01-level), and future extensions may adopt more advanced fusion schemes beyond fixed linear interpolation. We also tested repeated reuse of the same contextual dimension within a cascade (e.g., using C multiple times), but observed only marginal gains, so we focused on non-redundant C/I/S combinations in the main text. The best F1 score on the development set was reported in Tables~\ref{tab:full_fusion}. The baseline (no Astigmatism module) achieved: Explicit bias F1=0.4467, Implicit bias F1=0.4375, emotional intensity F1=0.4211.

\subsection{LLM Prompt Template}
\label{app:llm_stability}

All LLMs were evaluated with the fixed prompt template in Table~\ref{tab:prompt}.
The placeholder \texttt{\{text\}} was replaced with each input sentence from the ChLGBT test set.

\subsection{Few-shot LLM Comparison}
\label{app:few_shot_llm}

Table~\ref{tab:few_shot_combined} extends the zero-shot comparison by evaluating the same LLM set under 1-, 5-, and 10-shot prompting. The MAAM scores shown in each task block are the same MacBERT-based reference results reported in the main experiments, included here only to provide a stable non-LLM comparison point.

\begin{table*}[t]
\centering
\scriptsize
\resizebox{\textwidth}{!}{%
\begin{tabular}{llcccccc}
\toprule
\textbf{Task} & \textbf{System} & \textbf{1-acc} & \textbf{1-f1} & \textbf{5-acc} & \textbf{5-f1} & \textbf{10-acc} & \textbf{10-f1} \\
\midrule
\multicolumn{8}{l}{\textbf{Explicit} (MAAM Acc: 0.48$\pm$0.05 / F1: 0.46$\pm$0.05)} \\
Explicit & ChatGPT & \textbf{0.47$\pm$0.06} & \textbf{0.43$\pm$0.05} & \textbf{0.48$\pm$0.04} & \textbf{0.46$\pm$0.05} & \textbf{0.49$\pm$0.04} & \textbf{0.47$\pm$0.03} \\
Explicit & LLaMA & 0.36$\pm$0.04 & 0.36$\pm$0.03 & 0.38$\pm$0.03 & 0.36$\pm$0.02 & 0.38$\pm$0.02 & 0.38$\pm$0.03 \\
Explicit & ERNIE & 0.40$\pm$0.05 & 0.36$\pm$0.04 & 0.40$\pm$0.05 & 0.39$\pm$0.03 & 0.41$\pm$0.03 & 0.40$\pm$0.04 \\
Explicit & DeepSeek & 0.46$\pm$0.06 & \textbf{0.43$\pm$0.06} & 0.46$\pm$0.02 & 0.44$\pm$0.02 & 0.48$\pm$0.02 & 0.46$\pm$0.02 \\
Explicit & Qwen & 0.44$\pm$0.08 & 0.40$\pm$0.06 & 0.46$\pm$0.02 & 0.44$\pm$0.02 & 0.47$\pm$0.03 & 0.45$\pm$0.02 \\
\midrule
\multicolumn{8}{l}{\textbf{Implicit} (MAAM Acc: 0.44$\pm$0.01 / F1: 0.45$\pm$0.02)} \\
Implicit & ChatGPT & \textbf{0.43$\pm$0.03} & \textbf{0.43$\pm$0.02} & \textbf{0.46$\pm$0.02} & \textbf{0.45$\pm$0.02} & \textbf{0.44$\pm$0.02} & \textbf{0.45$\pm$0.01} \\
Implicit & LLaMA & 0.37$\pm$0.04 & 0.35$\pm$0.03 & 0.36$\pm$0.01 & 0.37$\pm$0.01 & 0.38$\pm$0.01 & 0.39$\pm$0.01 \\
Implicit & ERNIE & 0.40$\pm$0.02 & 0.41$\pm$0.02 & 0.41$\pm$0.02 & 0.41$\pm$0.02 & 0.41$\pm$0.01 & 0.43$\pm$0.02 \\
Implicit & DeepSeek & 0.41$\pm$0.01 & 0.42$\pm$0.01 & 0.42$\pm$0.02 & 0.42$\pm$0.02 & \textbf{0.44$\pm$0.02} & 0.44$\pm$0.02 \\
Implicit & Qwen & 0.38$\pm$0.02 & 0.37$\pm$0.02 & 0.39$\pm$0.02 & 0.40$\pm$0.02 & 0.41$\pm$0.01 & 0.40$\pm$0.01 \\
\midrule
\multicolumn{8}{l}{\textbf{Emotional} (MAAM Acc: 0.52$\pm$0.02 / F1: 0.47$\pm$0.01)} \\
Emotional & ChatGPT & 0.44$\pm$0.02 & 0.44$\pm$0.02 & 0.46$\pm$0.02 & 0.48$\pm$0.02 & 0.47$\pm$0.01 & 0.48$\pm$0.01 \\
Emotional & LLaMA & 0.40$\pm$0.03 & 0.39$\pm$0.03 & 0.40$\pm$0.01 & 0.41$\pm$0.01 & 0.43$\pm$0.01 & 0.45$\pm$0.01 \\
Emotional & ERNIE & \textbf{0.47$\pm$0.02} & \textbf{0.48$\pm$0.02} & \textbf{0.49$\pm$0.02} & \textbf{0.50$\pm$0.01} & \textbf{0.50$\pm$0.01} & \textbf{0.51$\pm$0.01} \\
Emotional & DeepSeek & 0.45$\pm$0.01 & 0.47$\pm$0.01 & 0.46$\pm$0.02 & 0.48$\pm$0.02 & 0.46$\pm$0.02 & 0.46$\pm$0.02 \\
Emotional & Qwen & 0.41$\pm$0.02 & 0.41$\pm$0.01 & 0.41$\pm$0.02 & 0.43$\pm$0.02 & 0.45$\pm$0.01 & 0.45$\pm$0.01 \\
\bottomrule
\end{tabular}%
}
\caption{Few-shot performance across explicit, implicit, and emotional tasks. LLM results are compared across 1-, 5-, and 10-shot settings; each task block reports the corresponding MAAM reference performance in parentheses.}
\label{tab:few_shot_combined}
\end{table*}

The few-shot results show that additional demonstrations generally improve LLM performance, especially from 1-shot to 5-shot, but the gains are uneven across models and dimensions. ChatGPT and DeepSeek are strongest on explicit and implicit bias, while ERNIE performs best on emotional intensity, suggesting that the three annotation dimensions favor different model behaviors. Even with few-shot prompting, however, no single LLM dominates all dimensions: MAAM remains competitive on explicit and implicit bias, whereas the strongest LLM results mainly appear in the emotional dimension. This supports the main-text interpretation that MAAM provides a compact and stable reference under a fixed protocol, while few-shot LLM prompting can be beneficial but remains sensitive to model choice and task dimension.

\subsection{Extended Case Analysis and Mechanism Interpretation}
\label{sec:appendix_case_study}

This appendix provides an extended case analysis to further illustrate the internal mechanisms of the MAAM framework, including detailed explanations of semantic weighting and contextual prior calibration.
This section complements the main text with additional interpretability evidence, rather than introducing new model components.

\paragraph{Local Feature Focusing via the Myopia Module}

We revisit the following sample from the ChLGBT corpus:
``Yuepan (fatty), it's quite interesting to see a marriage-fraud gay talking about integrity!''
(\begin{CJK*}{UTF8}{gkai}“月半，骗婚的同性恋谈诚信挺有趣的！”\end{CJK*}).

This sentence contained explicit identity references combined with ironic evaluative expressions, which often distracted standard encoder-based models.
Table~\ref{tab:weights_main} reported the token-level semantic weights produced by the Myopia module.

As shown, tokens directly associated with discriminatory meaning—such as ``marriage-fraud'' (\begin{CJK*}{UTF8}{gkai}骗婚\end{CJK*}) and ``gay'' (\begin{CJK*}{UTF8}{gkai}同性恋\end{CJK*})—received the highest weights.
In contrast, rhetorical or functional tokens (e.g., degree adverbs and punctuation) were assigned comparatively lower weights.

This distribution indicated that the Myopia module emphasized locally salient semantic cues while suppressing surrounding contextual noise.
Importantly, this process did not rely solely on hard keyword matching; instead, weights were assigned through task-specific linguistic levels and bounded contextual smoothing, allowing flexible responses to varied linguistic realizations.

\paragraph{Contextual Prior Calibration via the Astigmatism Module}
\begin{figure*}[t]
    \centering
    \includegraphics[width=\textwidth]{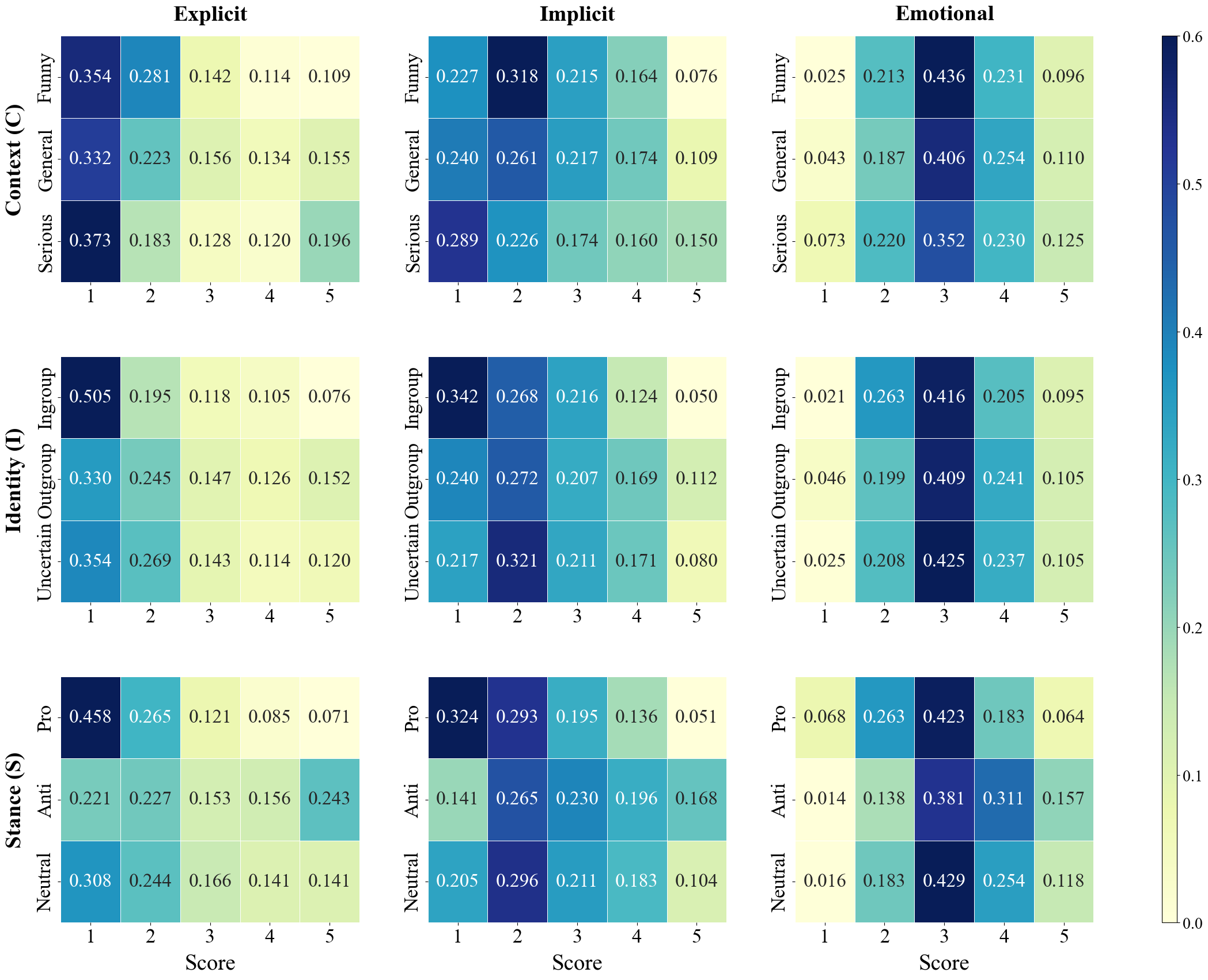}
    \caption{Heatmap of all dimension conditional score distribution for contextual prior calibration. }
    \label{fig:cis_heatmap}
\end{figure*}

After local feature extraction, MAAM applied the Astigmatism module to perform global calibration using the proposed C--I--S (Contextual Tone--Group Identity--Stance Polarity) framework.
For this example, the predicted contextual label combination was $(Funny, Outgroup, Anti)$.

In practical social media scenarios, the true speaker identity (Ingroup vs. Outgroup) was typically unobservable.
Rather than attempting explicit identity tracing, MAAM estimated contextual priors based on surface linguistic patterns and corpus-level statistics.
These estimates could be interpreted as maximum-likelihood approximations under partial observability.

Crucially, the C--I--S framework was incorporated as a soft prior rather than a hard decision rule.
Specifically, conditional probability distributions $P(\text{score} \mid \text{label})$ were constructed from training statistics and fused with model outputs through weighted integration.
As illustrated in Figure~\ref{fig:cis_heatmap}, these distributions exhibited clear separations across bias intensity levels.

This design ensured that moderate uncertainty or misclassification in individual contextual dimensions did not lead to unstable predictions.
Instead, the final output remained within a bounded confidence range, preventing disproportionate influence from any single contextual factor.

\paragraph{Portability and Modularity of the Framework}

The MAAM framework was designed to emphasize methodological generality rather than reliance on a specific encoder or dataset.
The process for constructing contextual prior tables—based on semantic clustering and frequency statistics—was model-agnostic and could be transferred to other bias domains, such as racial discrimination or workplace harassment.

Moreover, the framework supported modular decoupling.
Depending on task requirements and computational constraints, practitioners could activate only the Myopia module for local feature focusing, or selectively include subsets of the C--I--S dimensions within Astigmatism.
This flexibility allowed MAAM to operate across a range of deployment scenarios without structural modification.
\paragraph{Summary}

This extended case analysis provided detailed evidence for the dual-stage design of MAAM.
The Myopia module enhanced sensitivity to locally salient discriminatory cues, while the Astigmatism module stabilized predictions through contextual prior calibration.
Together, these components formed a robust and interpretable bias-aware modeling framework.
\end{document}